\documentclass[11pt]{article}

\usepackage[preprint]{acl}

\usepackage{times}
\usepackage{latexsym}
\usepackage{amssymb}
\usepackage{amsmath}
\usepackage{url}
\usepackage{algorithm}
\usepackage{algpseudocode}
\usepackage{graphicx} 
\usepackage{booktabs} 
\usepackage{array} 
\usepackage{multirow} 
\usepackage{graphicx} 
\usepackage[table]{xcolor} 
\usepackage{subfigure} 
\usepackage{subcaption}
\usepackage{graphicx}

\usepackage{booktabs}      
\usepackage{array}         
\usepackage{xcolor}        
\usepackage{arydshln}     
\usepackage{caption}       
\usepackage{graphicx}      
\usepackage{multirow}      
\usepackage{amsmath}
\usepackage{amssymb}
\usepackage{dsfont}
\usepackage{colortbl}
\usepackage{soul} 

\usepackage[T1]{fontenc}
\usepackage[utf8]{inputenc}
\usepackage{microtype}

\usepackage{inconsolata}

\usepackage{graphicx}

\newcommand{\replacecolor}[1]{{\textcolor[RGB]{181,68,106}{\textit{#1}}}}
\title{RTD-Guard: A Black-Box Textual Adversarial Detection Framework via Replacement Token Detection}

\author{He Zhu$^{*\dagger}$ , Yanshu Li$^{*\dagger}$ , Wen Liu$^{*\dagger}$, Haitian Yang$^{*}$   \\
  $^{*}$Institute of Information Engineering, Chinese Academy of Sciences, Beijing, China \\
  $^{\dagger}$School of Cyber Security, University of Chinese Academy of Sciences, Beijing, China \\}

\begin{document}
\maketitle
\begin{abstract}
Textual adversarial attacks pose a serious security threat to Natural Language Processing (NLP) systems by introducing imperceptible perturbations that mislead deep learning models. While adversarial example detection offers a lightweight alternative to robust training, existing methods typically rely on prior knowledge of attacks, white-box access to the victim model, or numerous queries, which severely limits their practical deployment. This paper introduces RTD-Guard, a novel black-box framework for detecting textual adversarial examples. Our key insight is that word-substitution perturbations in adversarial attacks closely resemble the “replaced tokens” that a Replaced Token Detection (RTD) discriminator is pre-trained to identify. Leveraging this, RTD-Guard employs an off-the-shelf RTD discriminator—without fine-tuning—to localize suspicious tokens, masks them, and detects adversarial examples by observing the prediction confidence shift of the victim model before and after intervention. The entire process requires no adversarial data, model tuning, or internal model access, and uses only two black-box queries. Comprehensive experiments on multiple benchmark datasets demonstrate that RTD-Guard effectively detects adversarial texts generated by diverse state-of-the-art attack methods. It surpasses existing detection baselines across multiple metrics, offering a highly efficient, practical, and resource-light defense mechanism—particularly suited for real-world deployment in resource-constrained or privacy-sensitive environments.
\end{abstract}

\section{Introduction}
With the widespread adoption of Pre-trained Language Models (PLMs)~\cite{devlin2019bert,radford2019language,DBLP:journals/corr/abs-1909-11942,liu2019roberta}across critical Natural Language Processing (NLP) applications, the robustness of these powerful systems has emerged as a primary security concern~\cite{zhao2025negatively}. Specifically, textual adversarial attacks—which manipulate model predictions through subtle, often human-imperceptible alterations to the input—reveal a significant fragility in current state-of-the-art architectures~\cite{vaswani2017attention}. These malicious inputs not only undermine the reliability of deployed services but also pose severe risks in safety-critical domains. Adversarial example detection offers a lightweight defense paradigm, identifying malicious inputs before they reach a deployed model, thus avoiding the high cost of robust training. However, the practical deployment of existing detection methods often relies on one or more unrealistic assumptions: 1) the need for known adversarial examples for training; 2) white-box access to model gradients or internal features; 3) a large number of queries to the victim model. These requirements fundamentally conflict with the real-world constraints of commercial NLP services—where model parameters are proprietary, attack patterns are unknown a priori, and low-latency response is essential.

An ideal practical detector should therefore simultaneously satisfy: strict black-box access (via API only), no prior knowledge of attacks, and minimal query overhead. Existing methods struggle to maintain high detection performance under this triple constraint. For instance, data-free black-box methods~\cite{radford2019language,mozes2020frequency}  exhibit limited generalization, while gradient-based or query-intensive methods face bottlenecks in access rights or efficiency.

Pre-trained Language Models serve as reservoirs of linguistic capabilities, which can be repurposed for diverse tasks with minimal adaptation~\cite{brown2020language, schick2021exploiting, gao2021making}. The novel insight of this work is that the fundamental operation of word substitution in textual adversarial attacks structurally mirrors the objective of the Replaced Token Detection (RTD) pre-training task~\cite{clark2020electra}. The RTD task trains a discriminator to identify tokens in a sentence that have been replaced, inherently making it skilled at detecting contextually unnatural word substitutions—the primary characteristic of  adversarial perturbations. Building on this insight, we propose RTD-Guard, a training-free, strictly black-box, and highly efficient adversarial example detection framework. RTD-Guard leverages a pre-trained RTD discriminator (used as-is, without fine-tuning) to localize ``suspiciously replaced'' tokens in an input text. It then intervenes by masking these tokens and detects adversarial examples by observing the consequent shift in the black-box victim model's prediction confidence. The entire process requires only two queries to the model and depends on neither adversarial data, model parameters, nor gradient information.

Our main contributions are summarized as follows:

\begin{itemize}
    \item We are the first to repurpose the intrinsic capability of the RTD pre-training task for adversarial detection, proposing RTD-Guard. It serves as a plug-and-play guard module, offering a novel solution for defense under strict black-box conditions with zero prior adversarial examples.
    \item Through systematic ablation studies, we validate that the RTD discriminator is superior to traditional gradient-based attribution methods for locating adversarial perturbations. We reveal that its strength lies in capturing contextual inconsistency rather than predictive importance, thereby avoiding signal failure caused by adversarial label shifts.
    \item We conduct extensive experiments on multiple text classification datasets against four state-of-the-art word-level attack methods. Results show that RTD-Guard significantly outperforms existing baselines in detection performance while maintaining the lowest query overhead and runtime, confirming its efficiency and robustness for practical deployment.
\end{itemize}

\section{Related Work}

The development of adversarial example detection in NLP has followed several distinct trajectories, each characterized by specific resource dependencies or assumptions that critically determine its practicality under real-world black-box constraints. To clearly situate our contribution, we categorize existing methods according to these requirements.

\textbf{Methods Requiring Adversarial Training Data.} A substantial body of work relies on access to labeled adversarial examples to train a dedicated detector. For instance, DISP~\cite{zhou2019learning} trains a token-level classifier on adversarial examples generated from clean data, incurring significant data preparation and computational costs. Similarly, ADFAR~\cite{bao2021defending} employs multi-task learning to augment the victim model with detection capabilities, requiring both clean and adversarial data along with white-box fine-tuning. TextShield~\cite{shen2023textshield} trains an LSTM classifier on handcrafted saliency features extracted from model outputs, again depending on adversarial data for supervision. These methods often struggle to generalize to unseen attack strategies, limiting their applicability in dynamic threat environments.

\textbf{Methods Requiring White-box Access.} Another line of work leverages internal model information—such as gradients or intermediate feature representations—which restricts their use to fully transparent model settings. For example, MLE~\cite{lee2018simple} and RDE~\cite{yoo2022detection} perform density estimation on the victim model's output features, necessitating white-box access to extract these representations. GradMask~\cite{moon2022gradmask} computes token saliency via gradient-based attribution and detects adversarial examples by masking important tokens and monitoring confidence shifts. Although effective, its reliance on model gradients prevents deployment with proprietary model APIs or in black-box scenarios.

\textbf{Methods Relying on Extensive Model Queries.} Several black-box detection approaches operate without training data or internal access but incur substantial query overhead. WDR~\cite{mosca2022suspicious} trains a classifier on confidence changes resulting from the iterative deletion of each word, requiring $\mathcal{O}(L)$ queries per example, where $L$ is the sequence length. VoteTRANS~\cite{nguyen2023votetrans} employs synonym substitution and hard-label voting, with a query complexity of $\mathcal{O}(L \cdot T)$. Such high query costs render these methods impractical for high-throughput or latency-sensitive services.

\textbf{Data-Free Black-box Methods.} A limited number of approaches aim to operate under minimal assumptions. FGWS~\cite{mozes2020frequency} replaces low-frequency words with their most frequent synonyms using static frequency lists; however, its performance degrades against attacks that do not depend on rare-word substitutions. The PPL method~\cite{radford2019language} uses the perplexity of an external language model (e.g., GPT-2~\cite{radford2019language}) as an anomaly score, requiring no access to the victim model. Yet, its detection consistency is often unstable, particularly when facing semantically-preserving attacks.

RTD-Guard distinguishes itself from all the aforementioned categories. It requires no adversarial training data, unlike data-dependent detectors; no white-box access to model parameters or gradients, unlike feature- or gradient-based methods; and only two black-box queries, unlike query-intensive schemes. Furthermore, rather than relying on static heuristics (e.g., FGWS) or external LMs (e.g., PPL), RTD-Guard directly leverages a pre-trained model's intrinsic capability to detect unnatural token substitutions—a capability inherently aligned with the operational mechanism of word-level adversarial attacks. This unique combination of minimal assumptions, low overhead, and inherent detection alignment positions RTD-Guard as a particularly practical and deployable solution for real-world adversarial defense.

\section{Research Objectives}

The core challenge in adversarial example detection lies in achieving high detection accuracy under strict practical constraints—operating as a black-box, requiring no prior adversarial examples, and maintaining minimal computational overhead. Existing methods often depend on at least one of these impractical resources, limiting their real-world applicability. To overcome this, our research aims to construct a novel, resource-efficient detection framework by repurposing the inherent capabilities of pre-trained language models. Our research objectives are as follows:

{\bf RO1}: Design and implement a novel detection framework, RTD-Guard, that operates under strict black-box conditions, requires zero adversarial training examples, and achieves constant, low query complexity, thereby fulfilling the key requirements for real-world deployment.

{\bf RO2}: Establish and formalize the theoretical connection between the Replaced Token Detection (RTD) pre-training objective and adversarial word-substitution attacks, providing a principled foundation that explains why an RTD discriminator is inherently effective for perturbation localization.

{\bf RO3}: Empirically validate the superiority of RTD-Guard through comprehensive experiments, demonstrating its state-of-the-art detection performance and computational efficiency across diverse datasets and attack methods, and systematically analyze its core design through ablation studies.

\section{Background}

\subsection{Preliminaries}

{\bf Adversarial examples} are carefully crafted inputs designed to deceive machine learning models while appearing normal to humans. In the text domain, these perturbations typically operate at three levels: character-level (e.g., typos), word-level (e.g., synonym substitution), and sentence-level (e.g., paraphrasing). Among these, word-level attacks strike an optimal balance between effectiveness and stealth, making them the most prevalent threat to production NLP systems.
 
{\bf Adversarial defense} strategies can be categorized into two main paradigms: robust training and adversarial detection. While robust training methods enhance model resilience through techniques like adversarial training, they often require significant computational resources and may compromise standard accuracy. Adversarial detection offers a complementary approach by operating as an independent module that identifies malicious inputs before they reach the target model.

{\bf Replaced Token Detection task (RTD)} first introduced with the ELECTRA architecture, offers a distinct pre-training paradigm. Unlike BERT's Masked Language Modeling (MLM), which trains a single model to predict original masked tokens, RTD employs a generator-discriminator framework. The generator is trained with a standard MLM loss to replace randomly masked tokens (typically 15\% of the sequence) with plausible alternatives:
\[
L_{\text{MLM}} = \mathbb{E} \left( -\sum_{i \in \mathcal{C}} \log p_{\theta_G}\left(\tilde{x}_{i, G}=x_i \mid \tilde{x}_G\right) \right),
\]
where $\tilde{x}_G$ is the input with randomly masked tokens at indices $\mathcal{C}$. The output of the generator is then used to construct a corrupted sequence $\tilde{x}_D$:
\[
\tilde{x}_{i, D}=\left\{\begin{array}{cc}
\tilde{x}_i \sim p_{\theta_G}\left(\tilde{x}_{i, G}=x_i \mid \tilde{x}_G\right), & i \in \mathcal{C} \\
x_i, & i \notin \mathcal{C}
\end{array}\right.
\label{eq:rtd_sample}
\]

This corrupted sequence serves as input to the discriminator, which is trained as a binary classifier to perform the RTD task: distinguishing between original tokens and generator-produced replacements. Its loss is:
\[
L_{\text{RTD}}=\mathbb{E}\left(-\sum_i \log p_{\theta_D}\left(\mathds{1}\left(\tilde{x}_{i, D}=x_i\right) \mid \tilde{x}_D, i\right)\right),
\]
where $\mathds{1}$ is the indicator function. The models are trained jointly with a combined loss $L = L_{\text{MLM}} + \lambda L_{\text{RTD}}$, where $\lambda$ heavily weights the discriminator's objective, thus making the discriminator a powerful, standalone detector of unnatural token substitutions.

\subsection{Problem Statement}

Consider a text classification model \(f: \mathcal{X} \rightarrow \mathcal{Y}\) and an input sequence \(x = [x_1, x_2, \dots, x_n]\). An adversary crafts an adversarial example \(x'\) by applying minimal perturbations such that \(f(x') \neq f(x)\), while \(x'\) remains semantically similar to \(x\). 

The goal of this work is to design a detector \(D: \mathcal{X} \rightarrow \{0,1\}\) that identifies whether an input \(x\) is adversarial (\(D(x)=1\)) or clean (\(D(x)=0\)).

\section{The RTD-Guard Framework}

\subsection{Threat Model}
\label{subsec:threat_model}
We define our defense scenario under a practical black-box threat model for  textual adversarial attacks.

\textbf{Attacker’s Goal \& Capabilities.} The attacker aims to craft an adversarial example \(x'\) from a clean input \(x\) such that the victim classifier \(f\) misclassifies it (\(f(x') \neq f(x)\)), while preserving semantic similarity to \(x\). We assume the attacker can:
\begin{itemize}
    \item Query the victim model \(f\) to obtain predictions and confidence scores.
    \item Employ state-of-the-art word-level attacks (e.g., TextFooler, PWWS, BAE) that rely on query-based search and synonym substitution.
    \item Has {no} access to the model's parameters, architecture, or training data.
\end{itemize}

\textbf{Defender’s Constraints}
The defender operates under the following strict and practical constraints:

\begin{itemize}
    \item \textbf{No white-box model access:} The victim model cannot be accessed for its internal parameters or gradient information (unlike {GradMask}, {MLE}, or {RDE}, which rely on gradients or internal features).
    \item \textbf{No prior adversarial examples:} The detector must be trained without any labeled adversarial data (unlike {DISP}, {ADFAR}, or {TextShield}, which require adversarial examples for supervision).
    \item \textbf{Minimal query overhead:} Only a very limited number of queries to the black-box model is permitted (unlike {WDR} or {VoteTRANS}, which require $\mathcal{O}(L)$ or $\mathcal{O}(L \cdot T)$ queries, where $L$ is the sequence length and $T$ is the number of transformations).
\end{itemize}

Methods such as {FGWS} and {PPL} align with these constraints, as they do not require white-box access, adversarial training data, or high query complexity. However, {FGWS} relies on static heuristic rules (e.g., word frequency substitution), and {PPL} leverages external language models (e.g., perplexity-based scoring), making their design fundamentally different from that of RTD-Guard.

RTD-Guard strictly adheres to the above constraints, focusing on adversarial detection without white-box model access, adversarial data, or excessive queries. Its design is tailored to operate efficiently under these realistic deployment conditions.

\subsection{Intuition and Motivation: Bridging RTD and Adversarial Perturbations}
\label{subsec:intuition}

\begin{figure*}[h]
\centering
\includegraphics[width=0.95\textwidth]{ 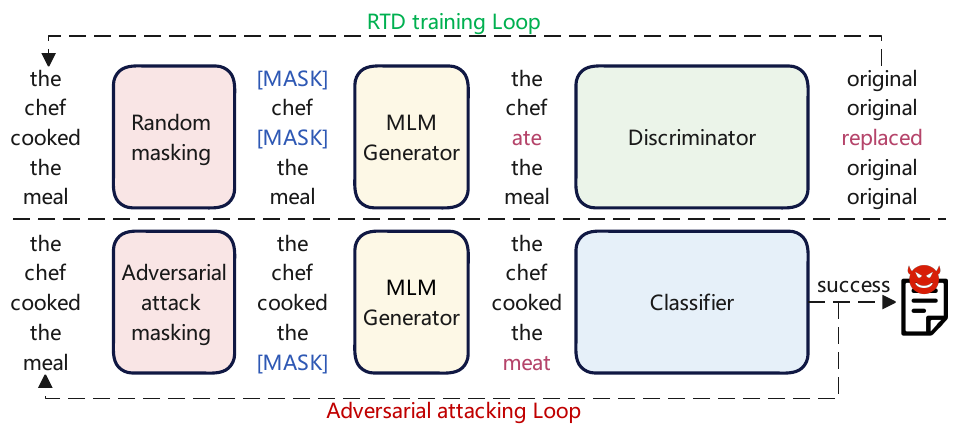}
\caption{Structural symmetry between adversarial attacks and RTD training. Both processes rely on token substitution as the core operation, but differ in their objectives. Adversarial attackers target critical tokens to mislead models, while RTD generators perform random substitutions to train discriminators.}
\label{fig:adv_rtd_symmetry}
\end{figure*}

The design of RTD-Guard is motivated by three fundamental properties of textual adversarial examples and pre-trained language models, which together bridge the gap between adversarial defense and a classic pre-training objective.

\textbf{Hypothesis 1: Fragility of Adversarial Examples.} Successful adversarial examples are inherently fragile. Crafting them typically requires an iterative, query-based search for “vulnerable” tokens and their optimal substitutions. Consequently, they reside in unstable regions of the input space where minor, targeted interventions can neutralize their malicious effect. Furthermore, to preserve semantics, attacks often push examples just beyond the model’s decision boundary. This critical state makes adversarial predictions highly sensitive to perturbations at precisely modified positions.

\textbf{Hypothesis 2: Adversarial Perturbations as Coordinated Optimization.} Effective word-level attacks are the product of optimization, not random noise. Due to complex feature interactions in deep models, misleading a classifier usually requires coordinated substitutions across multiple tokens. Thus, adversarial examples consist of a set of jointly optimized tokens engineered to flip the prediction while maintaining semantic coherence.

\textbf{Hypothesis 3: Model Robustness to Incidental Noise.} Deep learning models are generally robust to weak, random input variations—a resilience fostered by regularization techniques like dropout. Hence, randomly masking or perturbing tokens in a clean example typically induces only minor fluctuations in the model’s output confidence.

A pivotal insight follows: adversarial examples are highly sensitive to intervention at their optimized tokens, whereas clean examples remain stable. This differential sensitivity underpins our detection strategy: by accurately locating and neutralizing the pivotal adversarial tokens, we can induce a dramatic confidence shift for malicious inputs while leaving benign ones largely unaffected.

Traditional detection often mimics the attacker by using Word Importance Ranking (WIR) to find vulnerable tokens. This approach suffers from a fundamental mismatch: while the attacker uses the original true label $y$ to craft the example, the defender must detect based on the adversarially-induced label $y '$. This label shift severely degrades WIR-based detection.

We propose a more direct path: instead of reverse-engineering importance under a shifted label, we detect the artifacts of the substitution operation itself. This leads to our core conceptual link: the Replaced Token Detection (RTD) pre-training task. RTD is explicitly designed to identify contextually inconsistent tokens—those that appear to be substitutions—which is precisely the signature left by word-level adversarial attacks.

As illustrated in Figure \ref{fig:adv_rtd_symmetry}, a clear structural symmetry exists:
\begin{itemize}
\item The adversarial attacker (e.g., TextFooler) substitutes critical tokens with chosen synonyms to mislead the model.
\item The RTD generator substitutes random tokens with plausible alternatives to train the discriminator.
\end{itemize}

Both processes share the core operation of token substitution. Consequently, a discriminator excelling at RTD becomes inherently attuned to detecting “out-of-place” tokens—the hallmark of adversarial perturbations.

Our central thesis is that an RTD discriminator, by virtue of its pre-training, acquires a general, zero-shot capability to identify unnatural token substitutions. This capability transfers directly to adversarial perturbation detection, providing a model-agnostic and accurate localization mechanism. This insight obviates the need for adversarial training data, white-box access, or error-prone WIR estimation under label shift, establishing a principled and practical foundation for RTD-Guard.

\subsection{The RTD-Guard Framework}
\label{subsec:framework}

Building on the insight that a pre-trained RTD discriminator can identify adversarially replaced tokens, we formalize RTD-Guard—a four-stage detection pipeline for black-box settings. Given a victim classifier \(f\) and an input sequence \(x = [x_1, x_2, \dots, x_n]\),RTD-Guard outputs a binary detection decision as follows.

\noindent
\textbf{Step 1: Perturbation Localization.}
We employ a pre-trained RTD-based discriminator \(D_{\text{RTD}}\) to compute a {replacement probability} \(p_i \in [0,1]\) for each token \(x_i\):
\[
p_i = D_{\text{RTD}}(x, i).
\]
which reflects the token’s contextual inconsistency—a direct signal of potential adversarial substitution. The discriminator is used as-is, without any task-specific fine-tuning.

\noindent
\textbf{Step 2: Intervention via Masking.}
We select the top \(k\) tokens with the highest replacement probabilities. Let \(I = \text{TopK}(\{p_i\}_{i=1}^n, k)\) denote their indices. An intervened sequence \(x^m\) is constructed by replacing each token at position \(i \in I\) with a \([MASK]\) token:
\[
x^m_i = 
\begin{cases}
[MASK], & \text{if } i \in I \\
x_i, & \text{otherwise}.
\end{cases}
\]
Masking disrupts the adversarial influence carried by these suspicious tokens.

\noindent
\textbf{Step 3: Confidence Shift Measurement.}
The black-box victim model \(f\) is queried twice: once with the original input \(x\) and once with \(x^m\). Let the predicted class for the original input be \(c = \arg\max f(x)\). The detection score \(S_x\) is computed as the change in confidence for class \(c\):
\[
S_x = \text{Dis}\big(f(x)_c, f(x^m)_c\big),
\]
where \(\text{Dis}\) is a simple difference function. We use the squared difference \(\text{Dis}(a, b) = (a - b)^2\) in our implementation, which amplifies larger shifts. This score captures the model's sensitivity to the masked tokens: adversarial examples typically exhibit a more pronounced confidence shift.

\noindent
\textbf{Step 4: Detection Decision.}
The final decision is made by comparing \(S_x\) to a threshold \(\tau\) (empirically determined on a validation set):
\[
D(x) = 
\begin{cases}
1 \quad (\text{adversarial}), & \text{if } S_x > \tau \\
0 \quad (\text{clean}), & \text{otherwise}.
\end{cases}
\]

\begin{algorithm}[t]
\caption{RTD-Guard}
\label{alg:rtd-guard}
\begin{algorithmic}[1]
\Require Input sequence \(x\), victim model \(f\), RTD discriminator \(D_{\text{RTD}}\), budget \(k\), threshold \(\tau\)
\Ensure Detection label \(D(x) \in \{0,1\}\)
\State Compute \(p_i \gets D_{\text{RTD}}(x, i)\) for all tokens \(i\)
\State \(I \gets \text{Top-K}(\{p_i\}, k)\)
\State Initialize \(x^m \gets x\)
\For{each \(i \in I\)} 
    \State Replace \(x^m_i \gets [MASK]\)
\EndFor
\State Query \(f(x)\) and \(f(x^m)\)
\State \(c \gets \arg\max f(x)\)
\State $S_x \gets Dis(f(x)_c, f(x^m)_c)$
\If{\(S_x > \tau\)}
    \State \Return 1
\Else
    \State \Return 0
\EndIf
\end{algorithmic}
\end{algorithm}

RTD-Guard acts as a lightweight, plug-and-play shield that requires no modification to the deployed NLP system. It satisfies the strict defender constraints outlined in Section~\ref{subsec:threat_model}: (1) zero-shot and training-free—no adversarial data or fine-tuning; (2) strictly black-box—only prediction API access, no gradients or parameters; (3) computationally lean—exactly two model queries and one discriminator forward pass per example, ensuring low latency and scalability in production.

\section{Experimental Setup}

To ensure a fair and direct comparison with existing baseline methods, our experiments are conducted on the standardized textual adversarial benchmark provided by RDE \cite{yoo2022detection}. This benchmark is constructed using the \textsc{TextAttack} framework \cite{morris2020textattack}, applying four state‑of‑the‑art word‑level attack methods—\textbf{BAE} \cite{garg_bae_2020}, \textbf{PWWS} \cite{ren2019generating}, \textbf{TextFooler} \cite{jin_is_2020}, and \textbf{TF‑adj}\cite{morris2020reevaluating}—on three widely‑used text‑classification datasets: \textbf{IMDB}\citep{maas2011learning} , \textbf{AG‑News}\citep{zhang2015character} , and \textbf{Yelp} \citep{zhang2015character}. All attacks are performed against a fine‑tuned BERT model, ensuring that the evaluation reflects a realistic and challenging threat model consistent with prior work.

\subsection{Datasets and Victim Models}

\begin{table*}[th]
\caption{Statistics of evaluation datasets. “Test Examples” reports the number of original clean examples and the corresponding adversarial examples generated per attack.}
\centering
\resizebox{\textwidth}{!}{%
\begin{tabular}{lcccccc}
\toprule
    \multirow{2}{*}{\textbf{Dataset}} & \multirow{2}{*}{\textbf{Domain}} & \multirow{2}{*}{\textbf{Task}} & \multicolumn{1}{c}{\textbf{Classes}} & \multicolumn{1}{c}{\textbf{Median }} & \multicolumn{1}{c}{\textbf{Test Examples}} \\
    & & & \textbf{Number}& \textbf{Length}& \textbf{(Orig. / Gen.)} \\
\midrule
    IMDB\citep{maas2011learning} & Movie Reviews & Sentiment Classification & 2 & 161 & 25K / 10K \\
    AG-News\citep{zhang2015character} & News Topic & Topic Classification & 4 & 44 & 7.6K / 7.6K \\
    Yelp\citep{zhang2015character} & Restaurant Reviews & Sentiment Classification & 2 & 152 & 38K / 5K  \\
\bottomrule
\end{tabular}%
}
\label{table:c4x_benchmark}
\end{table*}

We evaluate on three public benchmarks spanning different domains and text lengths: \textbf{AG‑News} (4‑class news categorization), \textbf{IMDB} (binary movie review sentiment), and \textbf{Yelp} (binary restaurant‑review sentiment). Victim models are BERT‑based classifiers fine‑tuned on the respective training splits; we use the versions provided by \textsc{TextAttack} to align with the attack generation pipeline. Table \ref{table:c4x_benchmark} summarizes key statistics of each dataset.

\subsection{Adversarial Attacks}
To evaluate RTD-Guard against the most prevalent and threatening form of textual adversarial attack, we focus on word-level perturbations. This choice is deliberate: character-level attacks~\cite{ebrahimi_hotflip_2018,li_textbugger_2019,brefeld_generating_2020} (e.g., typos) are often easily filtered by spell-checkers, while sentence-level reconstructions~\cite{behjati2019universal,song2020universal} suffer from low success rates and poor semantic similarity. In contrast, word-level attacks optimally balance stealth, success rate, and semantic preservation, representing the primary practical threat to NLP systems.

Our evaluation employs four state‑of‑the‑art word‑level attack methods, all implemented in the \textsc{TextAttack} framework:

\begin{itemize}
\item \textbf{TextFooler}~\cite{jin_is_2020} generates adversarial examples by replacing important words with carefully selected synonyms.
\item \textbf{PWWS}~\cite{ren2019generating} weights synonym substitutions by combining word saliency and a synonym‑swap probability.
\item \textbf{BAE}~\cite{garg_bae_2020} leverages a masked language model to generate and insert contextually coherent replacement tokens.
\item \textbf{TF-adj}~\cite{morris2020reevaluating}refines TextFooler with additional fluency and grammar constraints to produce more natural adversarial examples.
\end{itemize}

\subsection{Baseline Detectors}
We compare RTD-Guard against the following representative and strong baselines in adversarial text detection:

\begin{itemize}
\item \textbf{PPL}~\cite{radford2019language}: Utilizes the perplexity score from GPT-2 as an anomaly measure for detection, relying solely on an external language model.
\item \textbf{MLE}~\cite{lee2018simple}: A white-box detector that performs Gaussian density estimation on the victim model's output features.
\item \textbf{FGWS}~\cite{mozes2020frequency}: A frequency-based method that replaces low-frequency words with their most common synonyms to neutralize perturbations.
\item \textbf{RDE}~\cite{yoo2022detection}: Estimates the probability density of hidden representations in a white-box setting to identify out-of-distribution adversarial examples.
\item \textbf{GradMask}~\cite{moon2022gradmask}: A white-box approach that masks tokens with high gradient-based saliency and monitors confidence change.
\item \textbf{WDR}~\cite{mosca2022suspicious}: A query-intensive black-box method that trains a classifier on word-level confidence shifts induced by token deletion.
\end{itemize}

These baselines collectively cover external‑LM scoring, static heuristics, white‑box feature analysis, and query‑based black‑box strategies, enabling a comprehensive assessment of RTD‑Guard under varied detection philosophies.

Note that while we include VoteTRANS~\cite{nguyen2023votetrans} in our efficiency analysis to demonstrate its computational characteristics, we exclude it from the main performance comparison. Its high query complexity ($O(L \cdot T)$) renders it computationally prohibitive for the large-scale evaluation across multiple datasets and attack methods conducted in this study.

\subsection{Evaluation Metrics}

We evaluate detection performance using three widely adopted metrics: the area under the receiver operating characteristic curve (\textbf{AUC}), which reflects overall separability between adversarial and clean examples; the \textbf{F1-Score}, which balances precision and recall; and the true positive rate at 10\% false positive rate (TPR10), which measures detection sensitivity under a low false-alarm constraint. For all metrics, higher values indicate better detection capability.

\subsection{Implementation Details}

We instantiate RTD-Guard using the pre-trained ELECTRA-large\cite{clark2020electra}\footnote{Model available at \url{https://huggingface.co/google/electra-large-discriminator}.}  discriminator as the RTD module without any fine-tuning. The top-\(k\) masking budget is set to \(k = 5\). The detection score is computed as the squared difference in prediction confidence, i.e., \(\operatorname{Dis}(a,b) = (a-b)^2\). Unless otherwise specified, we set \(\tau = 0.09\), which corresponds to a confidence shift of approximately \(0.3\). All runtime measurements are conducted on a workstation with an NVIDIA RTX 3090 GPU, and reported timings are averaged over 2,444 examples from the AG\textendash News/TextFooler subset.

\section{Experimental Results}

\newcommand{\GreenCell}[1]{\cellcolor[HTML]{D1F39D}#1}
\newcommand{\GreenRow}{\rowcolor[HTML]{D1F39D}}
\definecolor{captiongreen}{HTML}{D1F39D}

\sethlcolor{captiongreen}

\begin{table*}[t]
\centering
\caption{ Comparative Experimental Results of Text Adversarial example Detection. FGWS and PPL are highlighted in \hl{green} to denote that they share the same practical constraints as RTD‑Guard: strict black‑box access, no adversarial training data, and low query complexity.}
\resizebox{0.97\textwidth}{!}{%
\begin{tabular}{lcccccccccccc}
\hline\hline
  & \multicolumn{3}{c}{TextFooler} & \multicolumn{3}{c}{PWWS} & \multicolumn{3}{c}{BAE} & \multicolumn{3}{c}{TF-adj} \\ 
  \cmidrule(lr){2-4}\cmidrule(lr){5-7}\cmidrule(lr){8-10} \cmidrule(lr){11-13} 
 \multirow{-2}{*}{Methods} & TPR10& F1 & AUC & TPR10& F1 & AUC& TPR10& F1 & AUC&TPR10& F1 & AUC \\ \midrule 
\rowcolor{gray!20}{} & \multicolumn{12}{c}{\textbf{Ag-News}\;($|\mathbf{Y}|=4$ , Acc=0.925)} \\ 

\GreenRow
 PPL & 69.5& 77.5& 88.6& 66.2& 75.1& 86.9&23.0&34.7&66.3&31.4&44.4&65.9 \\
  MLE &78.3&83.2&93.5&71.6&78.9&91.8&68.3&76.8&91.2&73.5&80.2&90.8 \\
  \GreenRow
  FGWS &82.4&85.7&84.2&91.0&90.7&91.0&65.0&73.8&71.6&64.1&74.5&72.9 \\
 RDE &95.8&93.3&97.1&89.7&89.8&95.6&89.6&89.9&95.3&92.2&91.3&95.4 \\
  WDR &96.5&93.8&97.2&94.2&92.6&96.9&91.4&91.1&95.8&94.1&94.1&96.3 \\
   GradMask &95.4&93.1&96.4&90.9&91.5&95.6&90.6&91.1&94.5&99.0&94.0&93.7 \\

\GreenRow
{RTD-Guard} & 
\textbf{97.5} & \textbf{94.9} & \textbf{98.4} & \textbf{98.1} & \textbf{94.6} & \textbf{98.3} &
\textbf{98.1} & \textbf{94.8} & \textbf{97.8} &
\textbf{100} & \textbf{95.9} & \textbf{97.5} \\ 

\midrule
\rowcolor{gray!20} & \multicolumn{12}{c}{\textbf{IMDB}  \;($|\mathbf{Y}|=2$ , Acc=0.990)} \\ 
\GreenRow
 PPL &43.9&57.0&75.1&33.3&46.5&69.5&23.7&35.5&65.0&24.6&36.8&68.5 \\
 MLE &86.5&88.1&94.4&75.9&81.7&92.3&82.1&85.5&93.6&91.9&91.1&95.8  \\
 \GreenRow
 FGWS &84.6&87.1&87.3&88.2&89.1&90.8&62.1&72.3&70.9&72.6&80.6&78.4 \\
 RDE &96.6&93.5&96.6&88.2&89.0&94.5&93.1&91.7&95.6&\textbf{98.7}&\textbf{95.1}&\textbf{98.0}\\
 WDR &97.6&94.4&95.9&92.5&91.4&95.3&95.9&93.1&95.7&98.2&95.9&96.8 \\
 GradMask &95.7&93.4&96.0&91.3&91.7&95.6&92.3&92.0&95.2&96.2&93.8&95.9 \\

 \GreenRow
{RTD-Guard} & 
\textbf{98.5} & \textbf{94.8} & \textbf{98.3} & \textbf{98.1} & \textbf{93.0} & \textbf{98.0} &
\textbf{98.0} & \textbf{95.3} & \textbf{98.1} &
96.3 & 94.2 & 96.5 \\  \midrule
\rowcolor{gray!20} & \multicolumn{12}{c}{\textbf{Yelp} \;($|\mathbf{Y}|=2$ ,  Acc=0.983)} \\ 
\GreenRow
 PPL &39.8&53.1&76.8&37.6&51.0&74.8&17.8&27.8&66.5&13.9&22.5&62.7 \\
 MLE &33.8&47.1&65.1&41.2&54.5&66.3&41.2&54.5&66.3&27.8&40.4&66.0  \\
 \GreenRow
 FGWS &84.7&87.3&87.6&88.9&89.5&91.0&62.4&72.5&70.3&79.7&85.5&82.3 \\
 RDE &95.8&93.1&96.3&83.8&86.5&94.4&90.4&90.3&95.3&92.1&91.4&94.9\\
 WDR &98.2&95.6&97.5&94.5&92.7&96.7&96.6&94.8&97.2&\textbf{100}&\textbf{96.2}&96.6 \\
 GradMask &98.8&95.7&97.9&94.8&92.8&97.3&96.7&94.2&97.4&\textbf{100}&95.8&\textbf{98.2} \\

 \GreenRow
{RTD-Guard} & 
\textbf{99.1} & \textbf{95.8} & \textbf{98.5} & \textbf{97.6} & \textbf{95.2} & \textbf{98.6} &
\textbf{98.9} & \textbf{96.0} & \textbf{98.3} &
\textbf{100} & \textbf{97.1} & \textbf{98.9} \\ \bottomrule
\end{tabular}}

\label{tab:main_results}
\end{table*} 

\subsection{Main Results: Detection Performance}

As shown in Table~\ref{tab:main_results}, RTD-Guard consistently achieves superior or competitive performance across all three datasets and four attack methods. In 11 out of 12 dataset--attack combinations, it attains the highest scores in TPR10, F1-Score, and AUC, demonstrating robust and reliable detection capability.

Analysis of the baselines reveals distinct limitations. While \textbf{PPL} operates with minimal assumptions (no adversarial data or model access), its performance is highly unstable, deteriorating sharply against more semantically-constrained attacks such as BAE and TF-adj. White-box density estimators \textbf{MLE} and \textbf{RDE} exhibit sensitivity to certain attack patterns, as seen in their performance degradation under PWWS. The frequency-based method \textbf{FGWS} struggles notably against BAE-style insertions, with AUC dropping to approximately 70.9\%.

The stronger baselines, \textbf{WDR} and \textbf{GradMask}\footnote{For GradMask, $k=2$ yields the best results on TF‑adj; $k=1$ is used for all other settings.}, maintain stable detection rates (AUC $\geq$95\% in most scenarios). However, their practicality is limited: WDR incurs a linear query cost $\mathcal{O}(L)$, making it infeasible for real-time deployment, while GradMask requires white-box gradient access.

Notably, RTD-Guard exhibits none of these weaknesses. It remains robust against PWWS, effective on constrained attacks (BAE, TF-adj), and capable of handling insertion-based perturbations. This consistent performance stems from its attack-agnostic design, which does not rely on assumptions about the attack strategy or dataset distribution.

In summary, the results confirm that RTD-Guard provides a highly effective, practical, and resource-efficient defense against diverse word-level adversarial attacks.

\subsection{Computational Efficiency}
\label{subsec:efficiency}

\begin{figure}[t]
    \centering
    \includegraphics[width=0.9\linewidth]{ 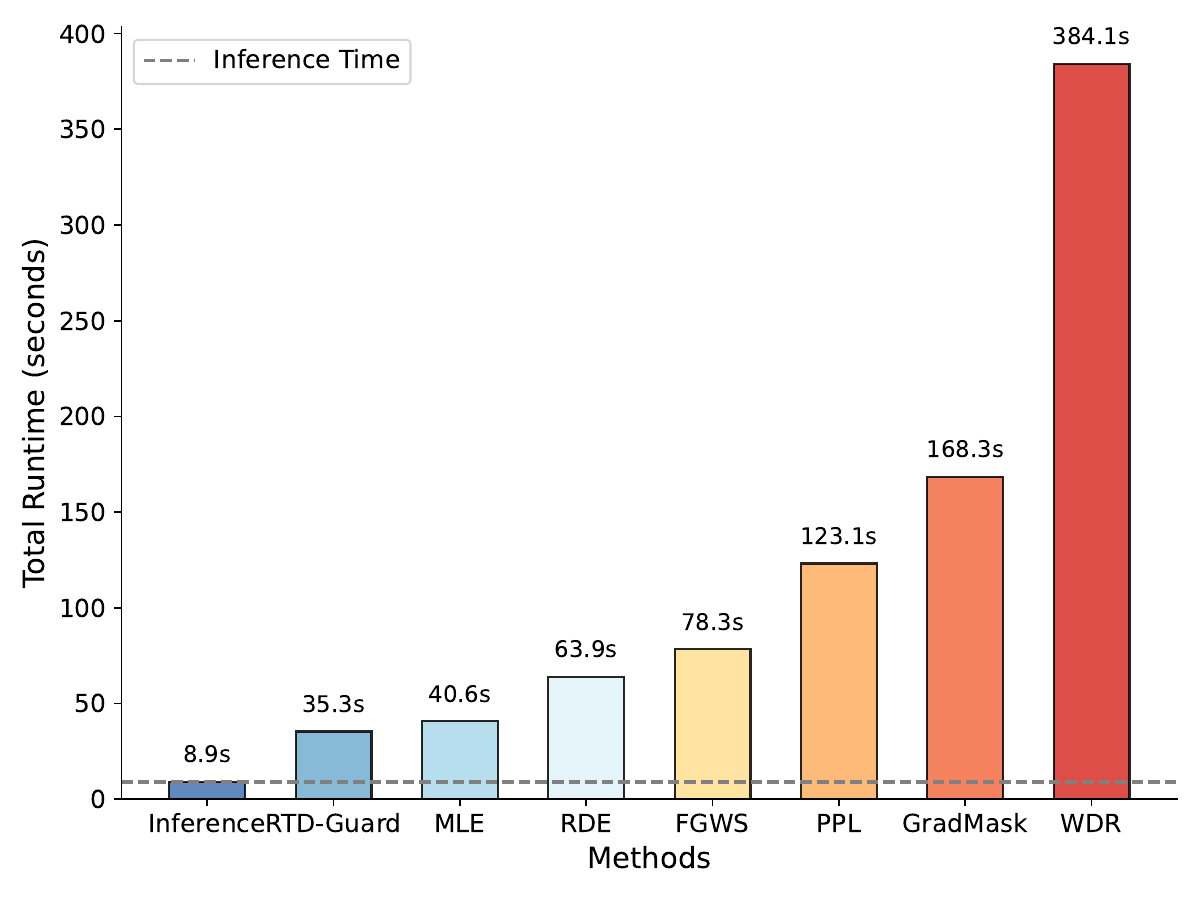}
    \caption{Total runtime comparison of different detection methods on the Ag-News/TextFooler split. RTD-Guard is the most efficient.}
    \label{fig:runtime_compare}
\end{figure}

Deployability in real-world scenarios critically depends on computational overhead. To assess efficiency, we measure the total execution time of each detection method on the AG‑News/TextFooler subset (2,444 examples). Results are summarized in Figure~\ref{fig:runtime_compare}.

\textbf{RTD‑Guard} achieves the lowest total runtime (35.3s), only marginally above the baseline inference time of the victim model alone (8.9s for 2,444 examples). This efficiency arises from its lightweight design: one forward pass through a frozen RTD discriminator plus exactly two black‑box queries to the victim model—yielding constant $O(1)$ query complexity.

In contrast, query-intensive methods incur substantial overhead. \textbf{WDR} requires $O(L)$ queries per example (where $L$ is sequence length), resulting in the longest runtime (384.1 seconds). \textbf{VoteTRANS} also suffers from high query cost ($O(L \cdot T)$), making it similarly impractical for latency-sensitive applications. While \textbf{GradMask} requires only two model accesses, its need to compute gradients in a white-box setting introduces significant computational burden (168.3 seconds).

Other lightweight baselines show varied efficiency. \textbf{PPL} relies on a large external LM (GPT-2) to compute perplexity, leading to moderate latency (123.1 seconds). \textbf{MLE} and \textbf{RDE}, as white-box feature-density methods, avoid repeated queries but still require internal feature extraction, resulting in runtimes of 40.6 and 63.9 seconds, respectively. \textbf{FGWS} performs static frequency-based substitution, which is efficient (78.3 seconds) but suffers from performance limitations as shown earlier.

In summary, RTD-Guard uniquely combines state-of-the-art detection performance with minimal computational overhead, offering a highly practical solution for real-time deployment in black-box settings.\footnote{All timing experiments were conducted on a workstation with a 3.8GHz 16-core AMD Ryzen 7 CPU, NVIDIA RTX 3090 GPU, and 64GB RAM running Windows 10.}

\subsection{Qualitative Analysis of Intervention Effects}

\label{subsec:qualitative_analysis}

\begin{table*}[t]
\caption{Performance comparison of different intervention strategies on AG-News.}
\label{tab:intervention_ablation}
\centering
\resizebox{0.95\textwidth}{!}{%

\begin{tabular}{lcccccccccccc}
\hline\hline
  & \multicolumn{3}{c}{TextFooler} & \multicolumn{3}{c}{PWWS} & \multicolumn{3}{c}{BAE} & \multicolumn{3}{c}{TF-adj} \\ 
  \cmidrule(lr){2-4}\cmidrule(lr){5-7}\cmidrule(lr){8-10} \cmidrule(lr){11-13} 
 \multirow{-2}{*}{Method} & TPR10& F1 & AUC & TPR10& F1 & AUC& TPR10& F1 & AUC&TPR10& F1 & AUC \\ \midrule 
\rowcolor{gray!20}{} & \multicolumn{12}{c}{\textbf{Ag-News}\;($|\mathbf{Y}|=4$ , Acc=0.925)} \\

 RTD-UNK &\textbf{97.8}&94.5&98.0&96.8&93.7&97.7  &95.7&94.3&97.3&\textbf{100}&95.3&96.3 \\
 RTD-DEL &97.6&94.5&97.6&96.2&93.8&97.4  &97.1&\textbf{94.8}&96.5&96.3&94.8&96.5 \\
RTD-MLM &97.5&94.8&98.0&97.4&94.0&97.9  &95.2&93.5&96.7&\textbf{100}&\textbf{96.4}&97.2 \\
 \GreenCell{RTD-Guard} & 
\GreenCell97.5 & \GreenCell\textbf{94.9} & \GreenCell\textbf{98.4} & \GreenCell\textbf{98.1} & \GreenCell\textbf{94.6} & \GreenCell\textbf{98.3} &
\GreenCell\textbf{98.1} & \GreenCell\textbf{94.8} & \GreenCell\textbf{97.8} &
\GreenCell\textbf{100} & \GreenCell95.9 & \GreenCell\textbf{97.5} \\ \hline
\bottomrule
\end{tabular}%
}
\end{table*}

Table~\ref{tab:ED_samples} provides a qualitative analysis of RTD-Guard's processing of adversarial inputs, illustrating its core operational behavior through representative examples. Three key patterns emerge from our analysis.

First, the RTD discriminator demonstrates a consistent ability to localize the core adversarial perturbations. Across diverse examples, the \textcolor[rgb]{0.00,0.00,1.00}{[MASK]} tokens in the RTD example are primarily the \replacecolor{substituted words} in the Adversarial example. This indicates that the pre-trained RTD objective successfully captures the contextual unnaturalness introduced by adversarial word-swaps, enabling precise targeting of the malicious modifications.

Second, the masking intervention effectively reverses the adversarial effect. As shown in the final column of Table~\ref{tab:ED_samples}, masking the suspicious tokens restores the model’s confidence in the original (correct) class to a level nearly matching that of the clean example. For instance, in the sports‑related example, confidence rebounds from 0.63 (Sci/Tech) to 0.98 (Sports). This sharp confidence shift upon masking RTD‑identified tokens directly serves as the detection signal and empirically validates our hypothesis that adversarial examples are fragile and can be neutralized by perturbing their critical tokens.

Third, RTD-Guard can localize dense perturbations involving multiple consecutive tokens (e.g., “fraud squad chief” → “hoax battalion leiter”) as well as sparse, scattered perturbations across the input. This capability arises from the RTD discriminator’s ability to independently evaluate the contextual consistency of each token.

We note that the lowercasing in the “RTD example” column results from the RTD discriminator’s tokenizer, which lowercases inputs during preprocessing. This normalization is confined to the detector and does not affect the victim model’s classification, as the original casing is preserved when the intervened sequence is passed to the victim model.

In summary, these qualitative observations corroborate the quantitative results, demonstrating that RTD-Guard’s core mechanism—locating contextually inconsistent tokens via a pre-trained RTD discriminator and neutralizing them via masking—reliably disrupts adversarial perturbations, offering a robust and interpretable detection approach.



\section{Ablation Studies and Analysis}
We conduct ablation studies to analyze the impact of core design choices in RTD-Guard, focusing on the following research questions (RQs):

\begin{itemize}
\item \textbf{RQ1:} How do different intervention strategies affect detection performance?
\item \textbf{RQ2:} How does token localization based on RTD compare to gradient-based attribution for adversarial detection?
\item \textbf{RQ3:} How does the scale of the RTD model affect detection performance?
\end{itemize}

\subsection{RQ1: Effect of Intervention Strategy}
\label{subsec:ablation_intervention}

Intervening on tokens identified as potentially adversarial is central to RTD-Guard. While our default method replaces them with a \texttt{[MASK]} token, other strategies can be applied. We compare four alternatives: replacing suspicious tokens with (1) an unknown token \texttt{[UNK]}, (2) deleting them entirely (DEL), and (3) using a Masked Language Model (MLM) to predict a contextually appropriate replacement. For MLM-Fill, we employ an RTD generator to predict the most contextually plausible token for each \texttt{[MASK]}.

Results on AG‑News (Table~\ref{tab:intervention_ablation}) show that detection performance remains consistently high across all intervention variants—RTD‑Guard (masking), RTD‑UNK, RTD‑DEL, and RTD‑MLM—with only minor fluctuations in TPR10, F1, and AUC scores. This finding directly supports our core hypothesis: the key to effective detection lies in the {accurate localization} of adversarial tokens, not the specific operation used to neutralize them. Once the pivotal tokens are correctly identified, any intervention that disrupts their influence—whether masking, deletion, or replacement—induces a pronounced confidence shift in the victim model. Hence, the simplest and most efficient strategy, masking with \texttt{[MASK]}, is both sufficient and optimal for practical deployment. Although RTD‑MLM yields a more natural‑looking intervened sequence, it does not improve detection, confirming that the primary advantage of our framework stems from its precise localization of adversarial substitutions.

\begin{figure*}[htbp]
    \centering
    \subfigure[RDE]{%
        \includegraphics[width=0.33\textwidth]{ 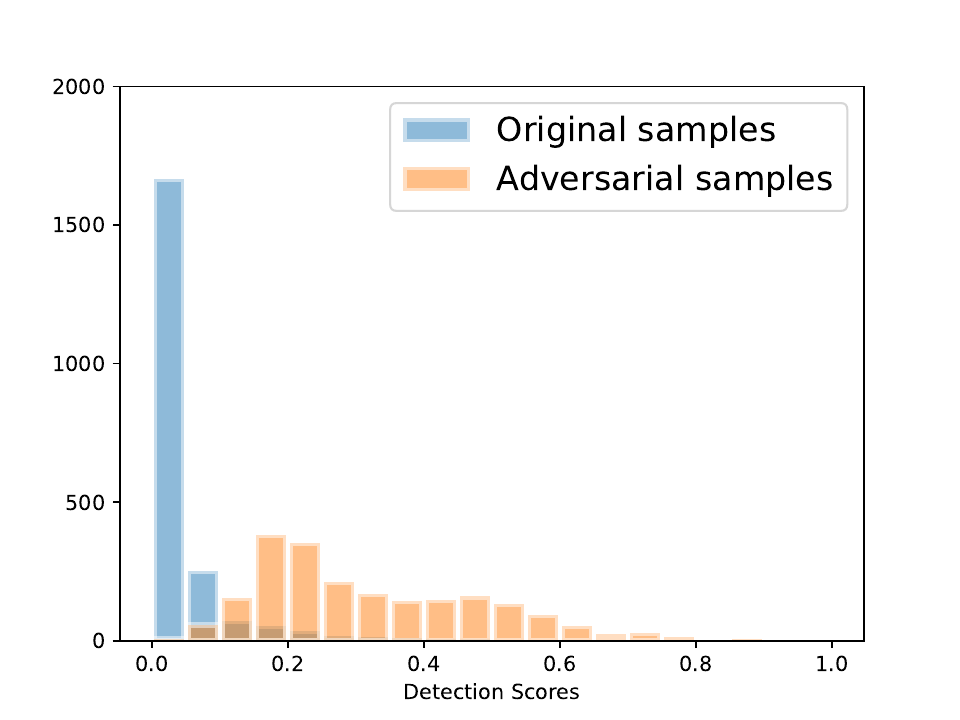}%
    }%
    \subfigure[GradMask]{%
        \includegraphics[width=0.33\textwidth]{ 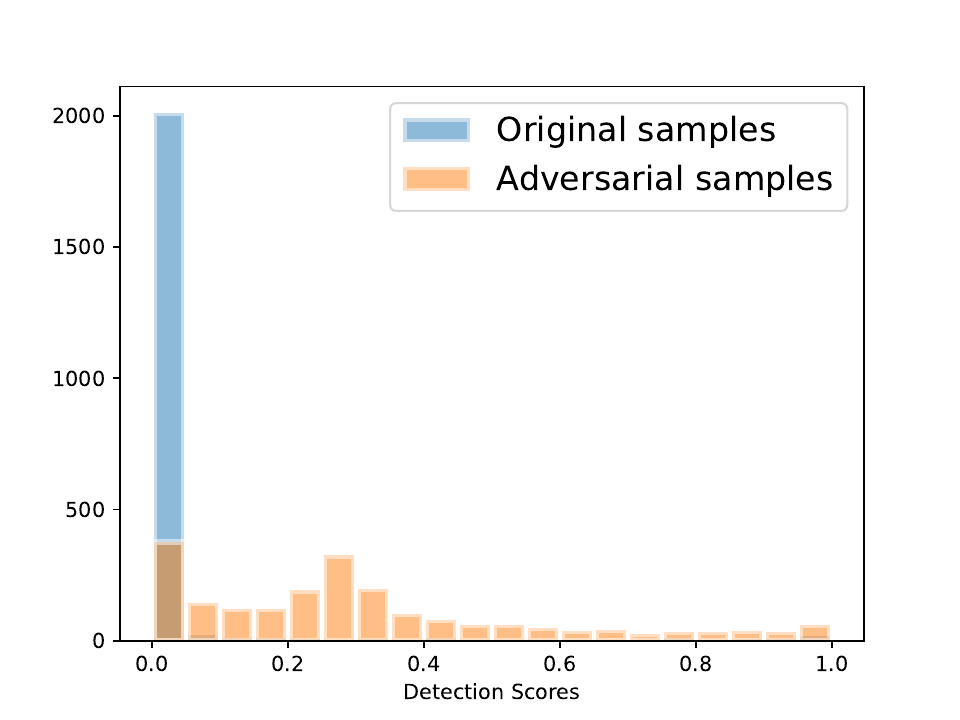}%
    }%
    \subfigure[RTD-Guard]{%
        \includegraphics[width=0.33\textwidth]{ 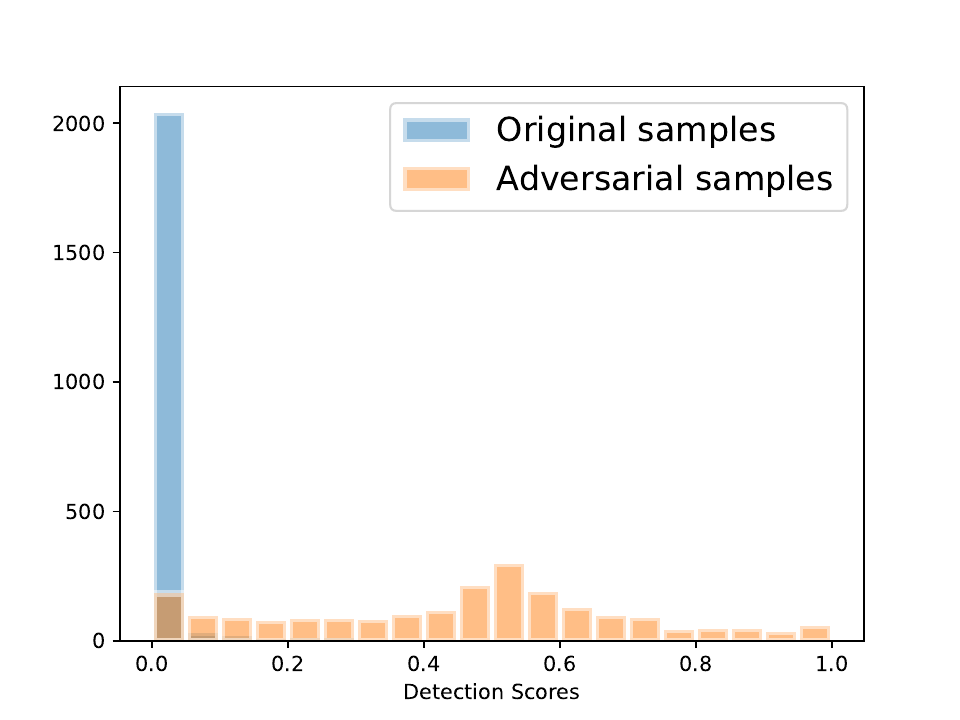}%
    }%
    \caption{Comparison of detection score distributions on the IMDB/TextFooler subset.}
    \label{fig:score_dist}
\end{figure*}

\subsection{RQ2: Divergence Between Gradient and RTD-based Token Localization}
\label{subsec:ablation_rtd_vs_grad}

A core strength of RTD-Guard lies in its ability to achieve state-of-the-art token localization using a pre-trained RTD discriminator, without relying on the victim model’s internal gradients. While gradient-based attribution remains one of the most compelling token-level signals—providing example-specific insights directly tied to the model’s internal optimization process—this signal is only accessible in white-box settings. To demonstrate the power of RTD-Guard and explore the fundamental nature of adversarial examples, we conduct a direct comparison with gradient-based methods that leverage privileged access to model gradients, as well as feature-based baselines.

We compare the detection score distributions of RTD-Guard, the gradient-based method GradMask, and the feature-density-based method RDE. As shown in Figure~\ref{fig:score_dist} (using IMDB/TextFooler), RTD-Guard produces scores with markedly less overlap and a sharper separation boundary between adversarial and clean examples. In contrast, both GradMask and RDE exhibit substantially greater overlap between the two classes, indicating weaker discriminability despite their access to internal model information.

\begin{figure*}[htbp] 
    \centering
    \subfigure[\tiny Ag-News/TextFooler]{%
        \includegraphics[width=0.24\textwidth]{ 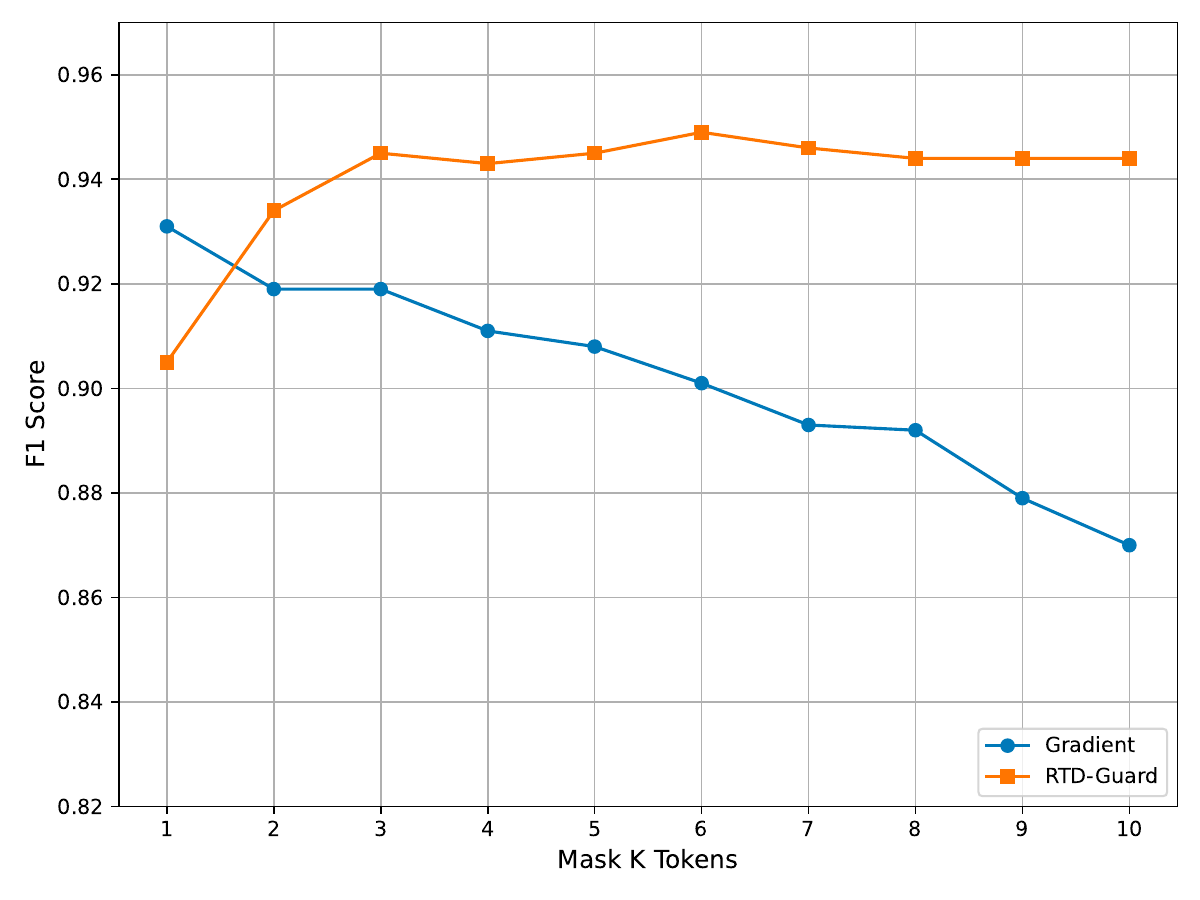}%
    }%
    \subfigure[\scriptsize Ag-News/PWWS]{%
        \includegraphics[width=0.24\textwidth]{ 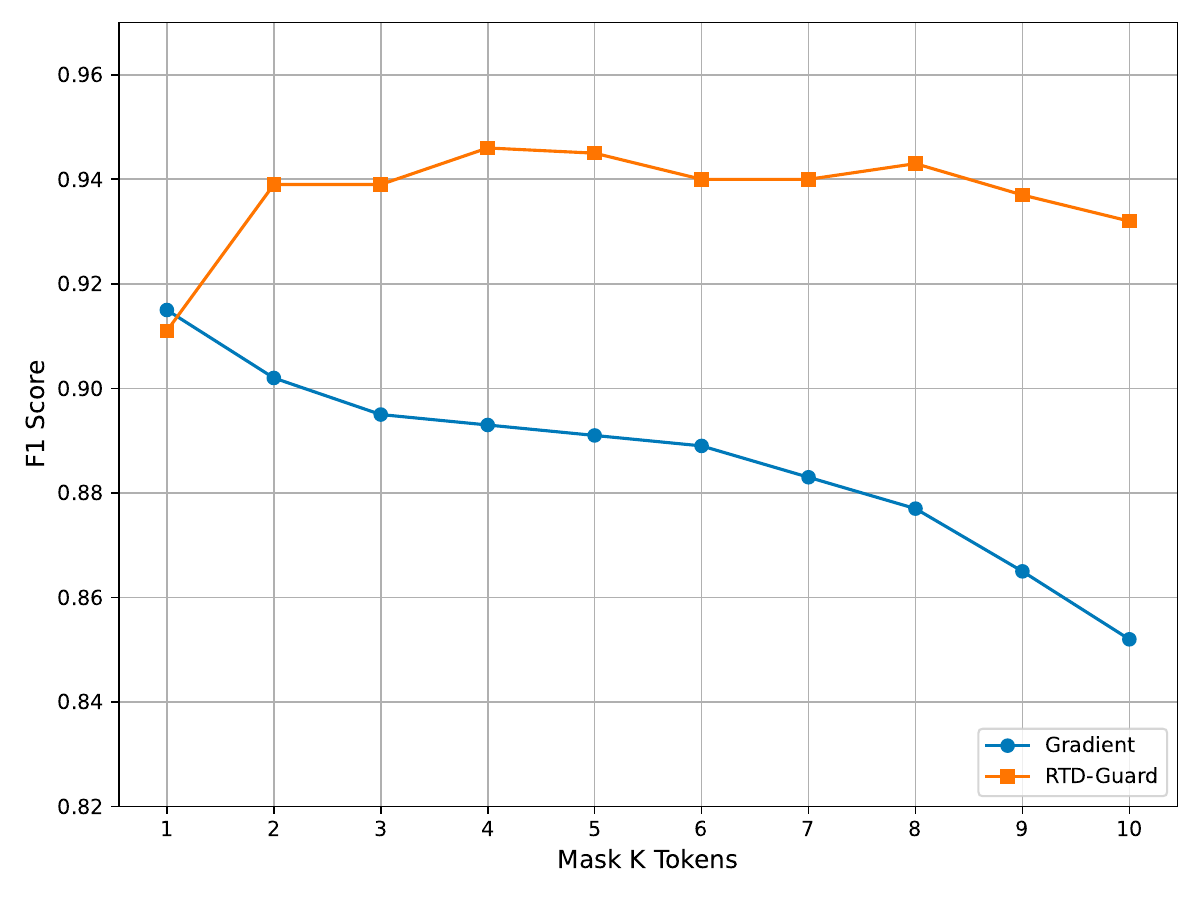}%
    }%
    \subfigure[\scriptsize Ag-News/BAE]{%
        \includegraphics[width=0.24\textwidth]{ 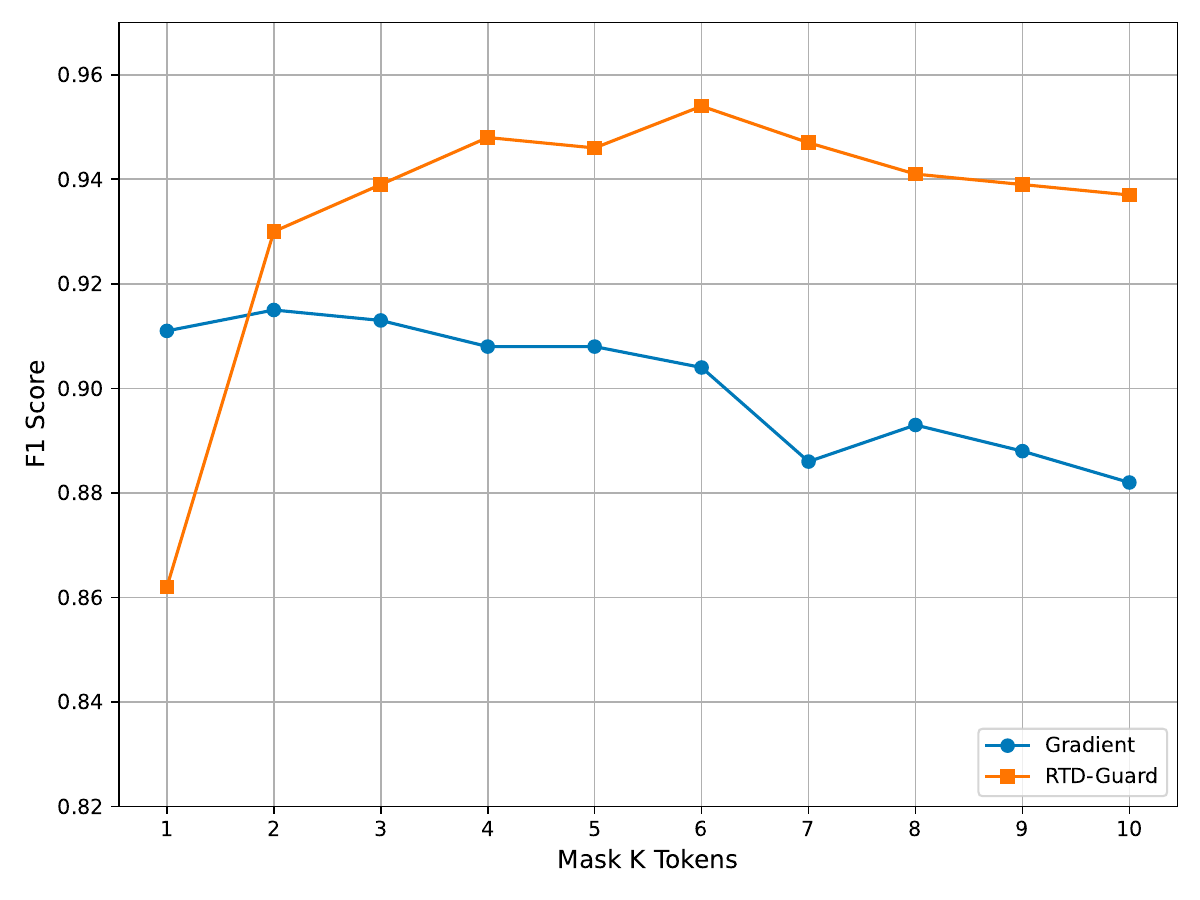}%
    }%
    \subfigure[\scriptsize Ag-News/TF-adj]{%
        \includegraphics[width=0.24\textwidth]{ 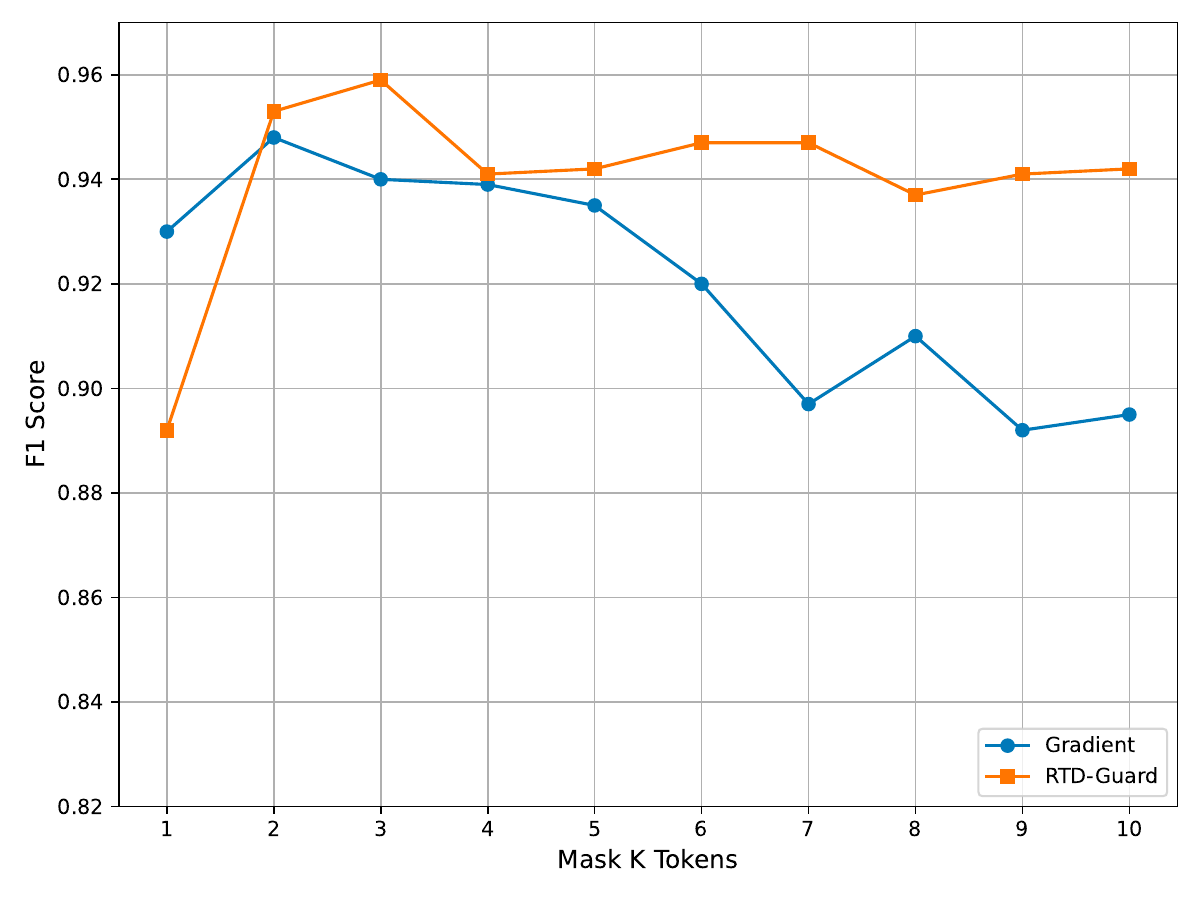}%
    }%


    \subfigure[\scriptsize Yelp/TextFooler]{%
        \includegraphics[width=0.24\textwidth]{ 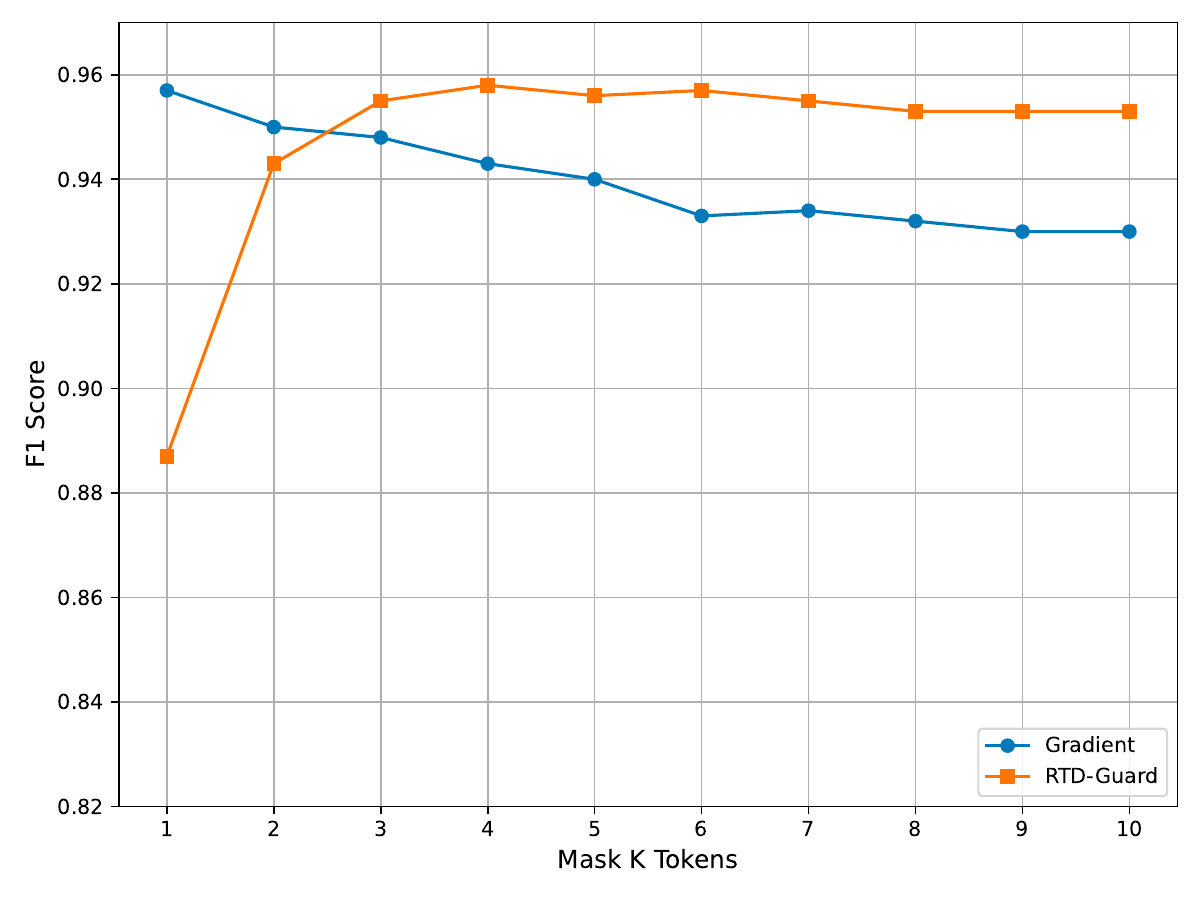}%
    }%
    \subfigure[\scriptsize Yelp/PWWS]{%
        \includegraphics[width=0.24\textwidth]{ 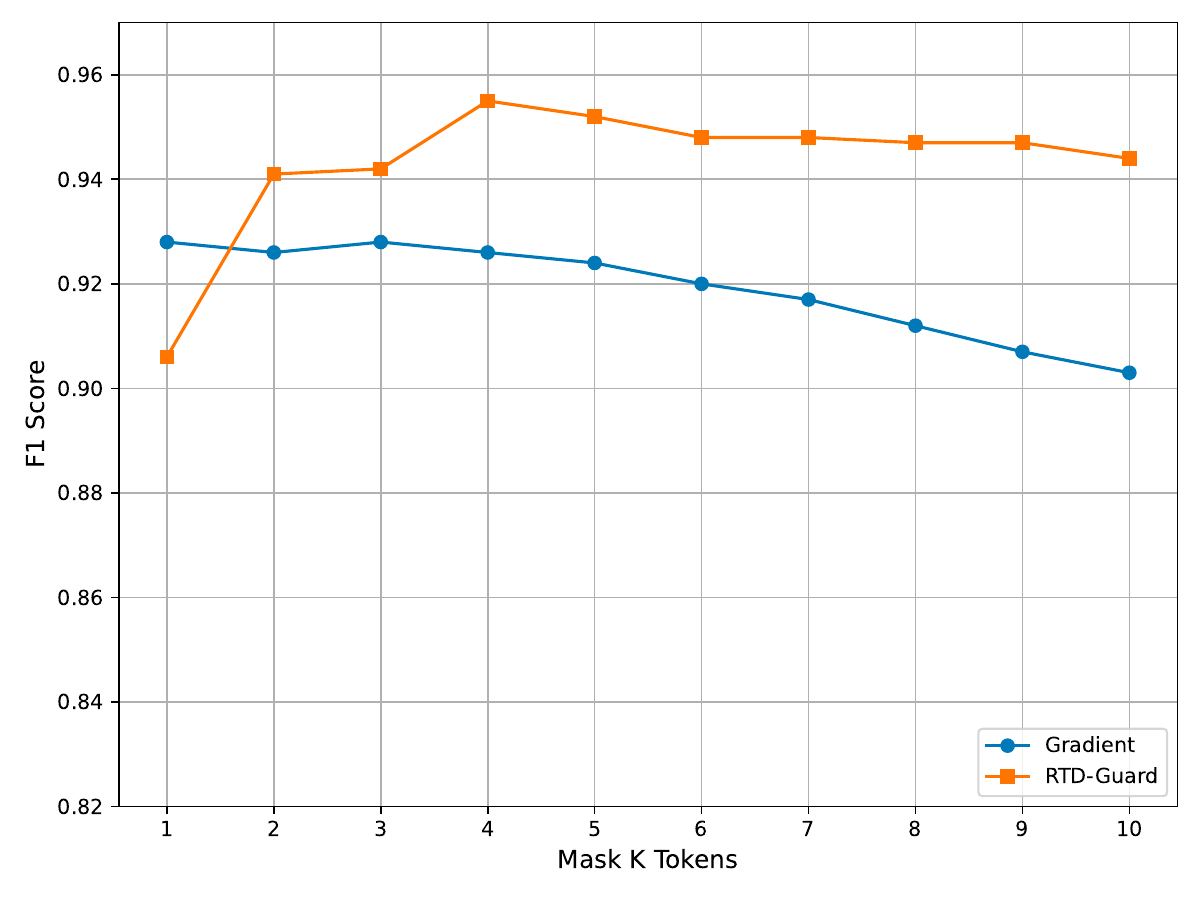}%
    }%
    \subfigure[\scriptsize Yelp/BAE]{%
        \includegraphics[width=0.24\textwidth]{ 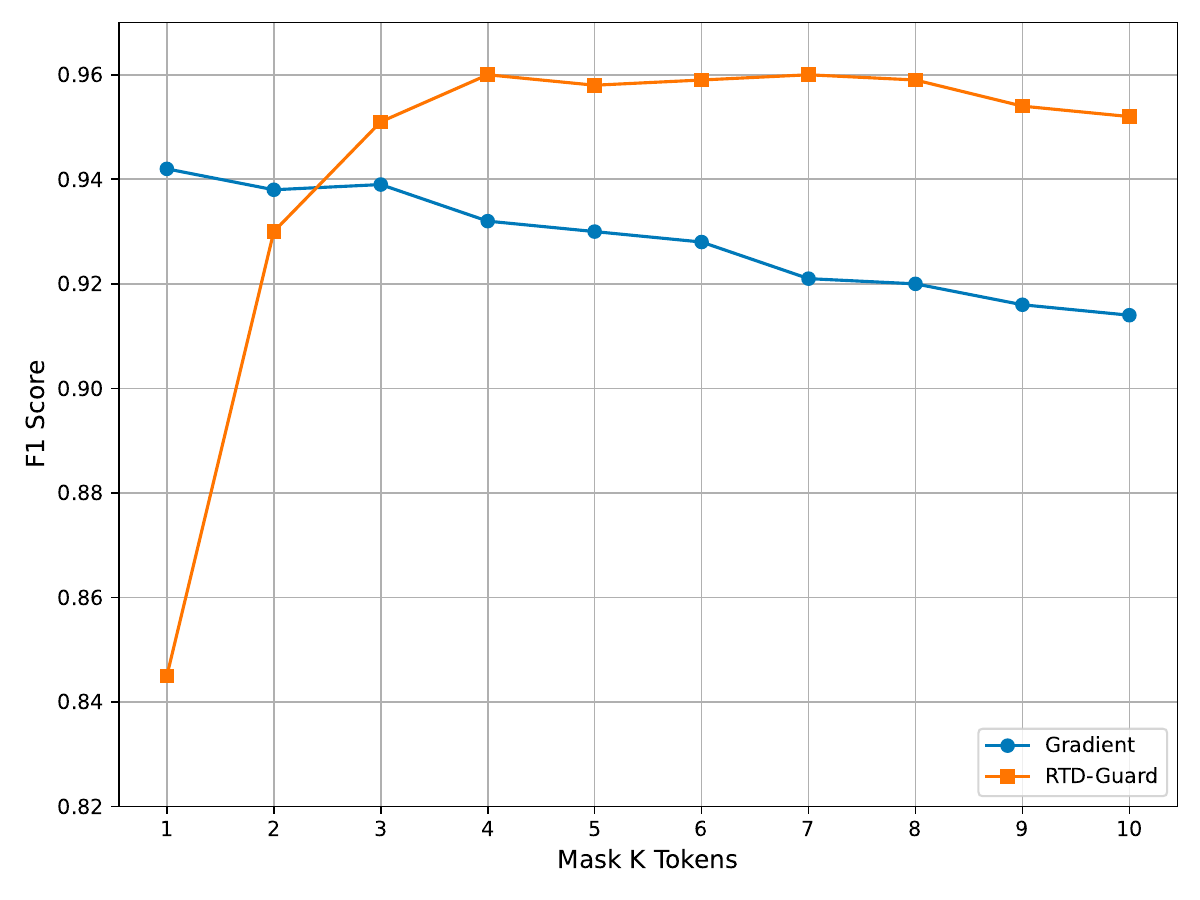}%
    }%
    \subfigure[\scriptsize Yelp/TF-adj]{%
        \includegraphics[width=0.24\textwidth]{ 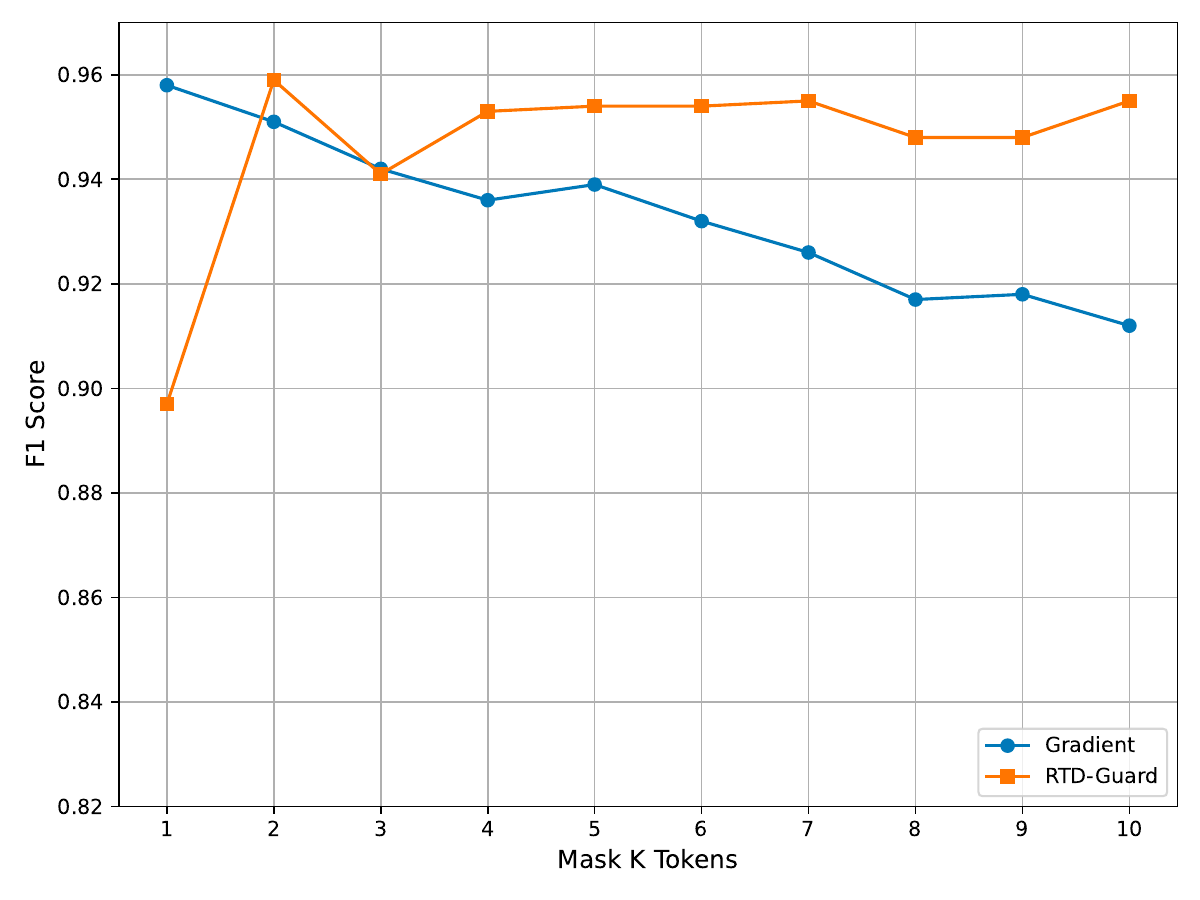}%
    }%


    \subfigure[\scriptsize IMDB/TextFooler]{%
        \includegraphics[width=0.24\textwidth]{ 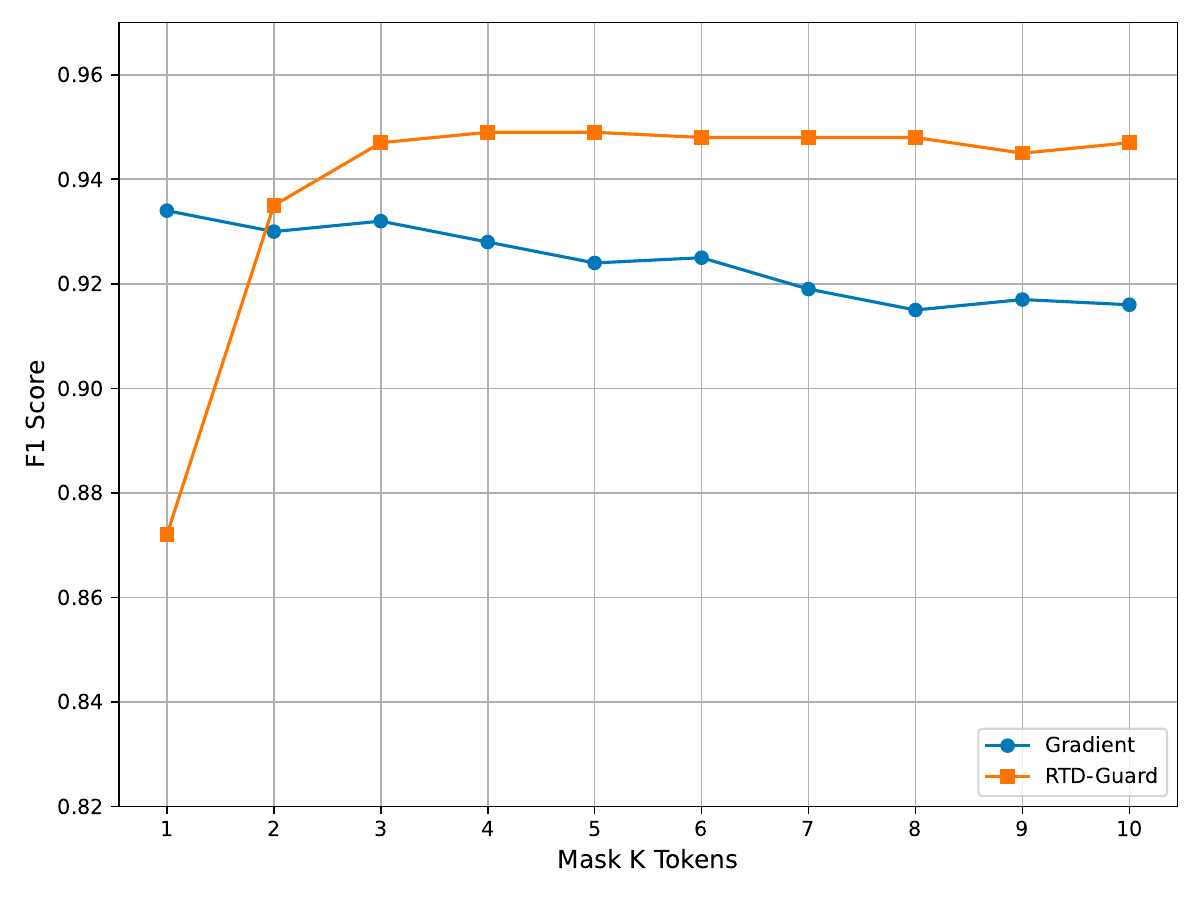}%
    }%
    \subfigure[\scriptsize IMDB/PWWS]{%
        \includegraphics[width=0.24\textwidth]{ 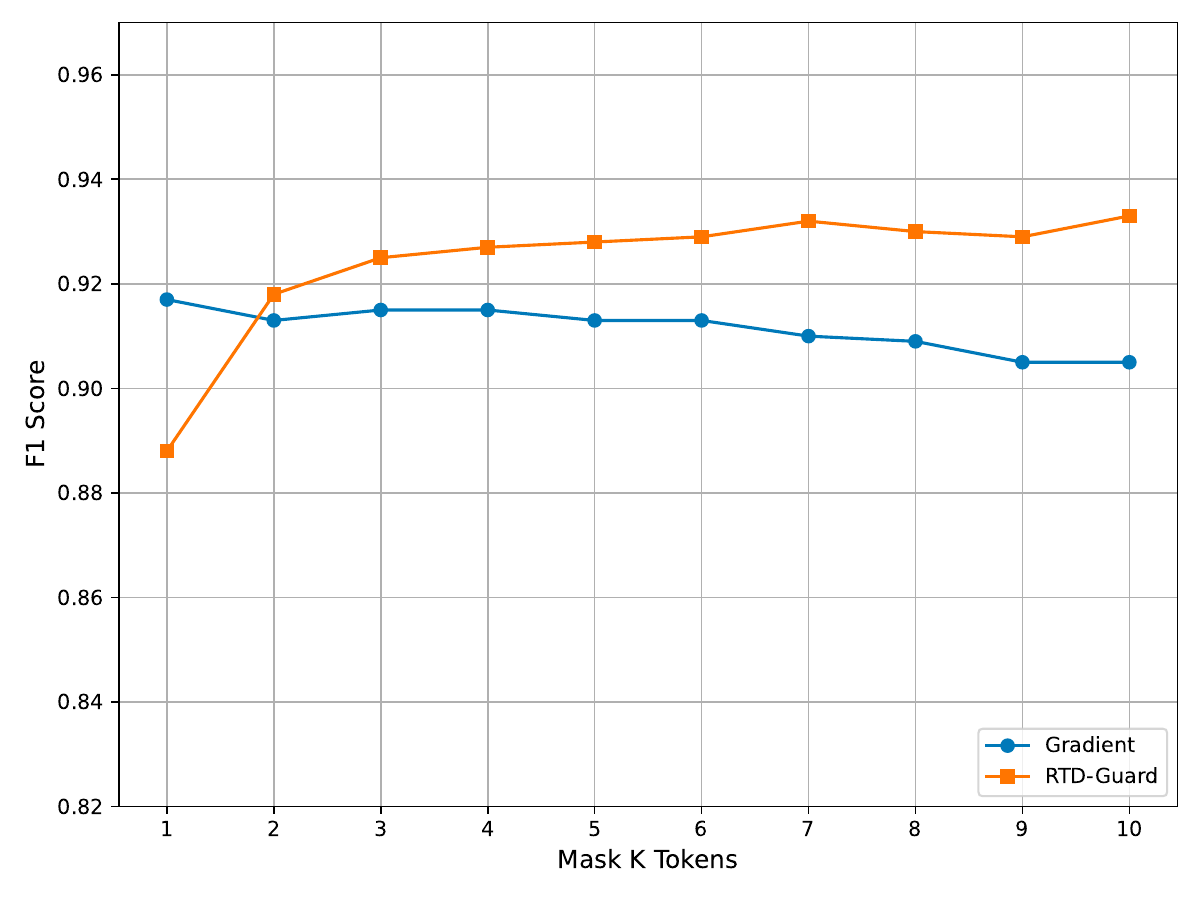}%
    }%
    \subfigure[\scriptsize IMDB/BAE]{%
        \includegraphics[width=0.24\textwidth]{ 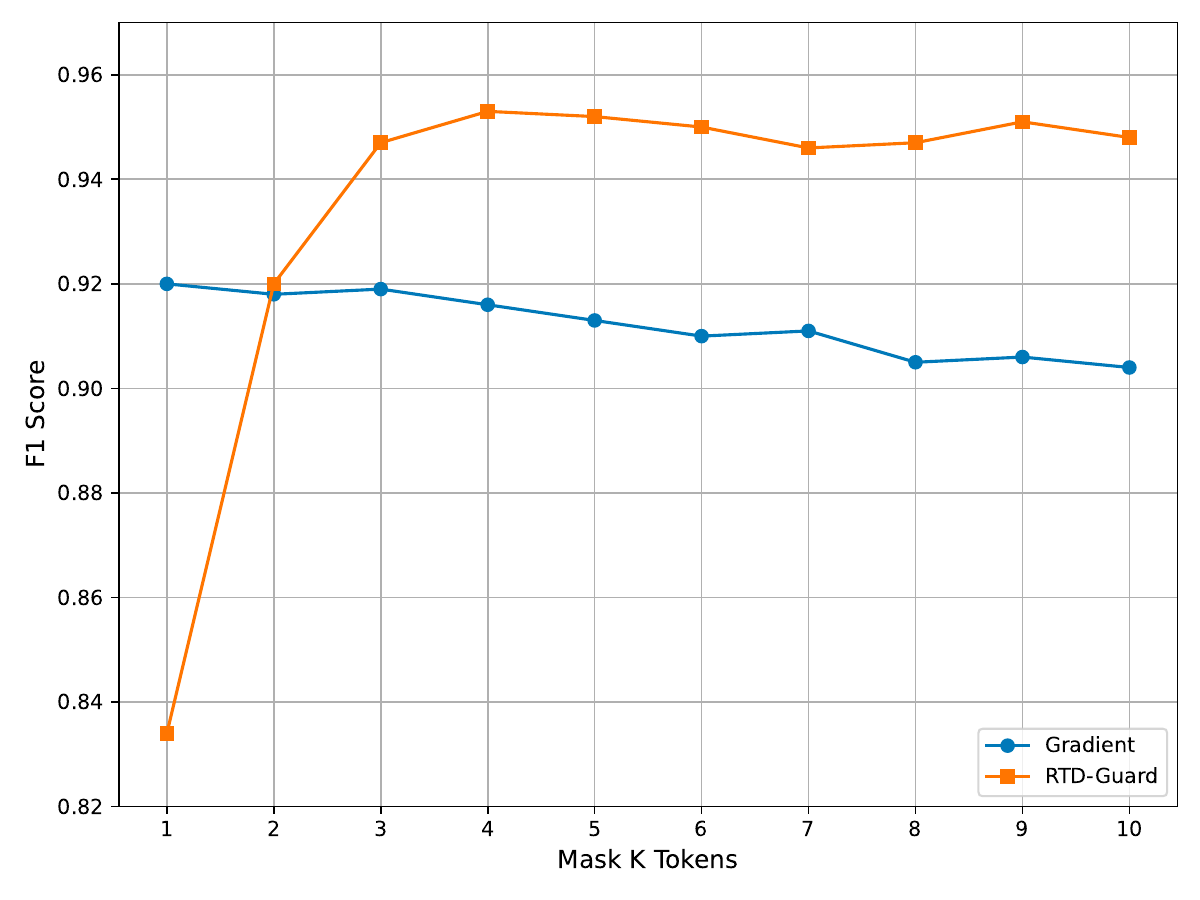}%
    }%
    \subfigure[\scriptsize IMDB/TF-adj]{%
        \includegraphics[width=0.24\textwidth]{ 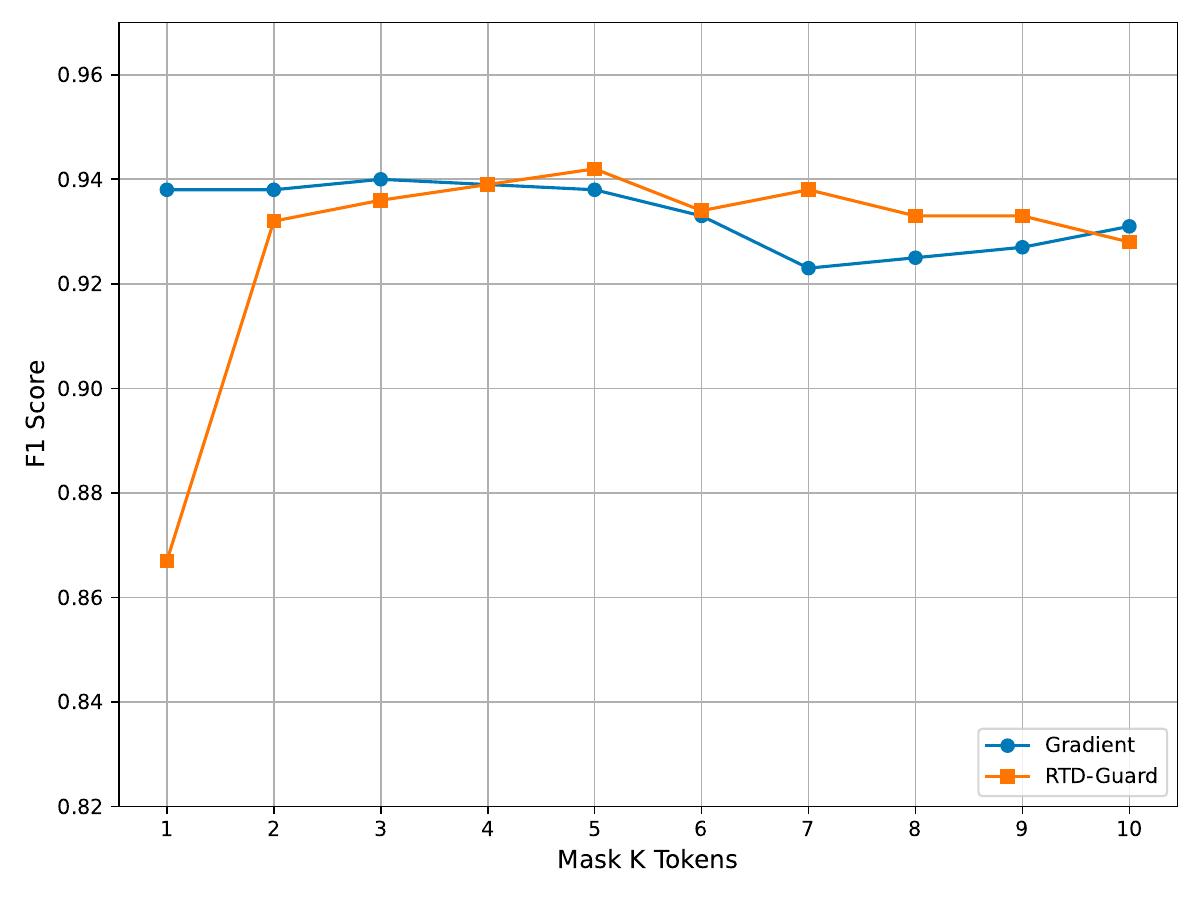}%
    }%

    \caption{Top-k masking performance comparison experiment across datasets and attack methods.}
    \label{fig:all_topk_compare} 
\end{figure*}

We further examine how detection performance responds to the masking budget \(k\). Sophisticated attacks often perturb multiple tokens; a robust defense should therefore neutralize several suspicious positions without harming clean accuracy. As \(k\) increases, we observe a sharp divergence: GradMask suffers a notable drop in overall F1-score—primarily due to collapsing clean accuracy—while RTD-Guard maintains stable or even improved performance. This contrast raises a key question: Why does raising \(k\) weaken gradient-based localization but strengthen RTD-based localization?

The root cause lies in what we term signal entanglement. Gradient-based attribution selects tokens that are important for the current prediction. In adversarial examples, these indeed correspond to the attack triggers; however, in clean examples, the same measure highlights semantic anchors—words that carry core meaning (e.g., sentiment adjectives). When \(k\) grows, GradMask inevitably masks such anchors in clean inputs, mutilating sentence semantics and triggering a surge in false positives. In short, gradients cannot disentangle adversarial influence from genuine semantic importance.

RTD-Guard operates on a different principle: it ranks tokens by their contextual inconsistency, a property largely orthogonal to semantic salience. Adversarial edits introduce syntactic-semantic irregularities that consistently rise to the top of the RTD ranking; meanwhile, the “least natural” tokens in clean sentences are typically proper nouns, rare entities, or syntactic connectors—not the semantic core. Increasing \(k\) thus allows RTD-Guard to progressively strip away adversarial noise without eroding semantic content, effectively performing surgical de-noising rather than semantic degradation.

\subsection{RQ3: Impact of RTD Model Scale on Detection Performance}

We further examine how the scale of the RTD discriminator affects detection capability. Although RTD pre-training has been adopted in various architectures (e.g., CoCo‑LM \cite{meng2021coco}, XLM‑E \cite{chi2021xlm}, CodeBERT \cite{fengcodebert}), publicly available RTD-based models remain predominantly from the ELECTRA family \cite{clark2020electra}. We therefore evaluate three standard ELECTRA variants: Small (14M parameters), Base (110M), and Large (335M).

As shown in Table \ref{tab:electra_size_abl}, while the 14M Small model already achieves successful detection, larger models deliver progressively stronger performance across all attacks and datasets.

Inference efficiency also scales with model size. On the AG‑News/TextFooler split, total processing times are 28.8s (Small), 33.1s (Base), and 35.3s (Large). After subtracting the fixed overhead of two victim‑model queries (2×8.9s), the net detection times are 11.0s, 15.3s, and 17.5s, respectively. These results indicate a clear trade‑off between accuracy and latency, allowing practitioners to select an appropriate RTD model size based on their specific deployment constraints.

\begin{table*}[t]
\centering

\caption{Result of RTD model scale ablation experiment}

\resizebox{0.97\textwidth}{!}{%
\begin{tabular}{lcccccccccccc}
\hline\hline
  & \multicolumn{3}{c}{TextFooler} & \multicolumn{3}{c}{PWWS} & \multicolumn{3}{c}{BAE} & \multicolumn{3}{c}{TF-adj} \\ 
  \cmidrule(lr){2-4}\cmidrule(lr){5-7}\cmidrule(lr){8-10} \cmidrule(lr){11-13} 
 \multirow{-2}{*}{Methods} & TPR10& F1 & AUC & TPR10& F1 & AUC& TPR10& F1 & AUC&TPR10& F1 & AUC \\ \midrule 
\rowcolor{gray!20}{} & \multicolumn{12}{c}{\textbf{Ag-News}\;($|\mathbf{Y}|=4$ , Acc=0.925)} \\ 

RTD-Small &96.1&93.4&97.7&95.4&93.3&97.5&90.9&92.9&95.2&93.9&94.2&95.1 \\
 RTD-Base &97.5&94.2&98.1&97.0&93.9&98.1&97.1&94.4&97.1&99.8&94.7&96.4 \\
 \GreenCell{RTD-Large} & 
\GreenCell\textbf{97.5} & \GreenCell\textbf{94.9} & \GreenCell\textbf{98.4} & \GreenCell\textbf{98.1} & \GreenCell\textbf{94.6} & \GreenCell\textbf{98.3} &
\GreenCell\textbf{98.1} & \GreenCell\textbf{94.8} & \GreenCell\textbf{97.8} &
\GreenCell\textbf{100} & \GreenCell\textbf{95.9} & \GreenCell\textbf{97.5} \\ \hline
\rowcolor{gray!20} & \multicolumn{12}{c}{\textbf{IMDB}  \;($|\mathbf{Y}|=2$ , Acc=0.990)} \\

 RTD-Small &97.5&94.7&97.8&92.8&91.8&97.0&94.4&93.0&96.4&95.8&93.8&96.7 \\
 RTD-Base &98.2&\textbf{95.1}&98.3&95.4&\textbf{93.4}&97.9&\textbf{98.6}&95.1&\textbf{98.4}&\textbf{97.4}&94.1&95.7 \\
 \GreenCell{RTD-Large} & 
\GreenCell\textbf{98.5} & \GreenCell94.8 & \GreenCell\textbf{98.3} & \GreenCell\textbf{98.1} & \GreenCell93.0 & \GreenCell\textbf{98.0} &
\GreenCell98.0 & \GreenCell\textbf{95.3} & \GreenCell98.1 &
\GreenCell96.3 & \GreenCell\textbf{94.2} & \GreenCell\textbf{96.5} \\ \hline

\rowcolor{gray!20} & \multicolumn{12}{c}{\textbf{Yelp} \;($|\mathbf{Y}|=2$ ,  Acc=0.983)} \\ 

 RTD-Small &98.5&95.6&98.4&96.3&94.2&98.2&97.3&94.3&96.9&100&\textbf{97.7}&97.4 \\
 RTD-Base &\textbf{99.3}&95.7&\textbf{98.7}&\textbf{97.9}&95.2&\textbf{98.9}&98.6&95.2&97.7&100&97.1&98.9 \\
 \GreenCell{RTD-Large} & 
\GreenCell99.1 & \GreenCell\textbf{95.8} & \GreenCell98.5 & \GreenCell97.6 & \GreenCell\textbf{95.2} & \GreenCell98.6 &
\GreenCell\textbf{98.9} & \GreenCell\textbf{96.0} & \GreenCell\textbf{98.3} &
\GreenCell\textbf{100} & \GreenCell97.1 & \GreenCell\textbf{98.9} \\ \hline
\end{tabular}}
\label{tab:electra_size_abl}
\end{table*}

\section{Implications}
\label{sec:implications}

\textbf{Theoretical Implications.}
Our work establishes a principled connection between adversarial text detection and a repurposed pre-training objective, offering a new lens through which to analyze and defend against word-level attacks. We demonstrate that adversarial word substitutions—far from being random noise—produce consistent {linguistic artifacts} characterized by contextual unnaturalness. These artifacts are detectable independently of the victim model’s architecture or training data, which explains RTD‑Guard’s strong generalization across diverse attacks and models.
Furthermore, we clarify why gradient‑based detection often falters: gradients reflect a token’s importance to the model’s {current} prediction, a signal that becomes unstable under the label shift induced by a successful attack. In contrast, RTD‑Guard relies on a model‑agnostic measure of contextual coherence, shifting the theoretical focus from modeling the victim’s decision boundary to directly characterizing the adversarial perturbation—a more stable and transferable foundation for detection.

\textbf{Practical Implications.}
These theoretical insights translate directly into deployable advantages. First, by detecting intrinsic linguistic irregularities, RTD‑Guard achieves robust performance without requiring knowledge of the attack algorithm or access to the victim model’s internals. This makes it a versatile, future‑proof module suitable for commercial APIs, edge devices, and other settings where models are proprietary and threats are unknown a priori.
Second, the framework is highly efficient by design: it requires exactly two black‑box queries and a single forward pass through a frozen discriminator, resulting in constant‑time overhead and low latency—critical for high‑throughput services.
Third, RTD‑Guard is both modular and interpretable. Its detection signal (confidence shift) and intervened examples provide actionable insights, easing integration and fostering operational trust.
Finally, by showing that general‑purpose pre‑training embeds latent {security capabilities}, this work points toward a broader paradigm: leveraging foundational linguistic competencies—beyond task‑specific fine‑tuning—to build more secure and resilient NLP systems.

\section{Conclusion}

We presented RTD-Guard, a black-box adversarial text detector that repurposes the Replaced Token Detection (RTD) objective into an efficient {training-free} defense. The core idea is simple yet effective: an RTD discriminator highlights contextually inconsistent tokens, and the detector then measures how the target model’s confidence changes when those suspicious tokens are masked. This design yields a practical detection pipeline that does not rely on adversarial training corpora, does not assume white-box access, and can be implemented with modest query budgets—properties that are often essential for deployment in real-world systems.

RTD-Guard offers a practical “plug-and-play” shield suitable for real-world deployment, especially in settings with strict black-box and resource constraints. Its design decouples defense from model architecture, ensuring scalability and low latency while preserving model transparency. Nevertheless, RTD-Guard also has an important limitation: it relies on an RTD model trained for a specific language to provide token-level inconsistency signals. As a result, the current detector does not yet generalize reliably across languages, and it may fail to effectively detect adversarial examples when applied to multilingual or cross-lingual inputs without language-matched RTD discriminators.

Future work can extend RTD-Guard along several directions with a particular focus on security challenges for large language models (LLMs): one promising avenue is to generalize the principle of detecting unnatural linguistic patterns via self-supervised inconsistency signals to prompt injection and jailbreak scenarios \cite{zou2023universal,liu2023autodan,liu2024autodan,lv2025hyper}, investigating how to identify hidden instructions, role hijacking, and privilege-escalation intents throughout instruction-following and tool-use pipelines and how to implement deployable, input-level risk warning and blocking mechanisms under black-box constraints; another important direction is hallucination detection, where RTD-based contextual consistency signals—coupled with response differences induced by counterfactual masking or replacement—may help surface factual inconsistencies, unsupported claims, or content with unverifiable provenance in model outputs, enabling lightweight hallucination detection and reliability estimation for generative systems.

\bibliography{custom}

@inproceedings{ebrahimi_hotflip_2018,
	address = {Melbourne, Australia},
	title = {{HotFlip}: {White}-{Box} {Adversarial} {Examples} for {Text} {Classification}},
	shorttitle = {{HotFlip}},
	url = {https://aclanthology.org/P18-2006},
	doi = {10.18653/v1/P18-2006},
	urldate = {2022-09-26},
	booktitle = {Proceedings of the 56th {Annual} {Meeting} of the {Association} for {Computational} {Linguistics} ({Volume} 2: {Short} {Papers})},
	publisher = {Association for Computational Linguistics},
	author = {Ebrahimi, Javid and Rao, Anyi and Lowd, Daniel and Dou, Dejing},
	year = {2018},
	pages = {31--36},
}

@inproceedings{behjati2019universal,
  title={Universal adversarial attacks on text classifiers},
  author={Behjati, Melika and Moosavi-Dezfooli, Seyed-Mohsen and Baghshah, Mahdieh Soleymani and Frossard, Pascal},
  booktitle={ICASSP 2019-2019 IEEE International Conference on Acoustics, Speech and Signal Processing (ICASSP)},
  pages={7345--7349},
  year={2019},
  organization={IEEE}
}

@inproceedings{li_textbugger_2019,
  title={TextBugger: Generating Adversarial Text Against Real-world Applications},
  author={Li, J and Ji, S and Du, T and Li, B and Wang, T},
  booktitle={26th Annual Network and Distributed System Security Symposium},
  year={2019}
}

@inproceedings{brefeld_generating_2020,
  title={Generating black-box adversarial examples for text classifiers using a deep reinforced model},
  author={Vijayaraghavan, Prashanth and Roy, Deb},
  booktitle={Joint European Conference on Machine Learning and Knowledge Discovery in Databases},
  pages={711--726},
  year={2019},
  organization={Springer}
}

@inproceedings{garg_bae_2020,
	address = {Online},
	title = {{BAE}: {BERT}-based {Adversarial} {Examples} for {Text} {Classification}},
	shorttitle = {{BAE}},
	url = {https://aclanthology.org/2020.emnlp-main.498},
	doi = {10.18653/v1/2020.emnlp-main.498},
	urldate = {2022-09-20},
	booktitle = {Proceedings of the 2020 {Conference} on {Empirical} {Methods} in {Natural} {Language} {Processing} ({EMNLP})},
	publisher = {Association for Computational Linguistics},
	author = {Garg, Siddhant and Ramakrishnan, Goutham},
	year = {2020},
	pages = {6174--6181},
}

@inproceedings{song2020universal,
  title={Universal adversarial attacks with natural triggers for text classification},
  author={Song, Liwei and Yu, Xinwei and Peng, Hsuan-Tung and Narasimhan, Karthik},
  booktitle={Proceedings of the 2021 Conference of the North American Chapter of the Association for Computational Linguistics: Human Language Technologies},
  pages={3724--3733},
  year={2021}
}

@inproceedings{morris2020textattack,
  title={TextAttack: A Framework for Adversarial Attacks, Data Augmentation, and Adversarial Training in NLP},
  author={Morris, John and Lifland, Eli and Yoo, Jin Yong and Grigsby, Jake and Jin, Di and Qi, Yanjun},
  booktitle={Proceedings of the 2020 Conference on Empirical Methods in Natural Language Processing: System Demonstrations},
  pages={119--126},
  year={2020}
}

@article{lee2018simple,
  title={A simple unified framework for detecting out-of-distribution samples and adversarial attacks},
  author={Lee, Kimin and Lee, Kibok and Lee, Honglak and Shin, Jinwoo},
  journal={Advances in neural information processing systems},
  volume={31},
  year={2018}
}

@inproceedings{zhou2019learning,
  title={Learning to discriminate perturbations for blocking adversarial attacks in text classification},
  author={Zhou, Yichao and Jiang, Jyun-Yu and Chang, Kai-Wei and Wang, Wei},
  booktitle={Proceedings of the 2019 Conference on Empirical Methods in Natural Language Processing and the 9th International Joint Conference on Natural Language Processing (EMNLP-IJCNLP)},
  pages={4904--4913},
  year={2019}
}

@inproceedings{mozes2020frequency,
  title={Frequency-guided word substitutions for detecting textual adversarial examples},
  author={Mozes, Maximilian and Stenetorp, Pontus and Kleinberg, Bennett and Griffin, Lewis},
  booktitle={Proceedings of the 16th conference of the European chapter of the association for computational linguistics: Main volume},
  pages={171--186},
  year={2021}
}

@inproceedings{yoo2022detection,
  title={Detection of adversarial examples in text classification: Benchmark and baseline via robust density estimation},
  author={Yoo, KiYoon and Kim, Jangho and Jang, Jiho and Kwak, Nojun},
  booktitle={Findings of the Association for Computational Linguistics: ACL 2022},
  pages={3656--3672},
  year={2022}
}

@inproceedings{mosca2022suspicious,
  title={“that is a suspicious reaction!”: Interpreting logits variation to detect NLP adversarial attacks},
  author={Mosca, Edoardo and Agarwal, Shreyash and Ram{\'\i}rez, Javier Rando and Groh, Georg},
  booktitle={Proceedings of the 60th Annual Meeting of the Association for Computational Linguistics (Volume 1: Long Papers)},
  pages={7806--7816},
  year={2022}
}

@inproceedings{moon2022gradmask,
  title={GradMask: Gradient-Guided Token Masking for Textual Adversarial Example Detection},
  author={Moon, Han Cheol and Joty, Shafiq and Chi, Xu},
  booktitle={Proceedings of the 28th ACM SIGKDD Conference on Knowledge Discovery and Data Mining},
  pages={3603--3613},
  year={2022}
}

@online{shen2023textshield,
  title={TextShield: Beyond Successfully Detecting Adversarial Sentences in Text Classification},
  author={Shen, Lingfeng and Zhang, Ze and Jiang, Haiyun and Chen, Ying},
  url={https://arxiv.org/abs/2302.02023},
  year={2023}
}

@inproceedings{zhang2015character,
  title={Character-level convolutional networks for text classification},
  author={Zhang, Xiang and Zhao, Junbo and LeCun, Yann},
  booktitle={Proc. of NeurIPS},
  year={2015}
}

@inproceedings{clark2020electra,
  title = {{ELECTRA}: Pre-training Text Encoders as Discriminators Rather Than Generators},
  author = {Kevin Clark and Minh-Thang Luong and Quoc V. Le and Christopher D. Manning},
  booktitle = {ICLR},
  year = {2020},
  url = {https://openreview.net/pdf?id=r1xMH1BtvB}
}

@online{DBLP:journals/corr/abs-1909-11942,
  author    = {Zhenzhong Lan and
               Mingda Chen and
               Sebastian Goodman and
               Kevin Gimpel and
               Piyush Sharma and
               Radu Soricut},
  title     = {{ALBERT:} {A} Lite {BERT} for Self-supervised Learning of Language
               Representations},
  journal   = {CoRR},
  volume    = {abs/1909.11942},
  year      = {2019},
  url       = {http://arxiv.org/abs/1909.11942},
  archivePrefix = {arXiv},
  eprint    = {1909.11942},
  bibsource = {dblp computer science bibliography, https://dblp.org}
}

@article{liu2019roberta,
  title={Roberta: A robustly optimized bert pretraining approach},
  author={Liu, Yinhan and Ott, Myle and Goyal, Naman and Du, Jingfei and Joshi, Mandar and Chen, Danqi and Levy, Omer and Lewis, Mike and Zettlemoyer, Luke and Stoyanov, Veselin},
  journal={arXiv preprint arXiv:1907.11692},
  year={2019}
}

@inproceedings{jin_is_2020,
  title={Is bert really robust? a strong baseline for natural language attack on text classification and entailment},
  author={Jin, Di and Jin, Zhijing and Zhou, Joey Tianyi and Szolovits, Peter},
  booktitle={Proceedings of the AAAI conference on artificial intelligence},
  volume={34},
  number={05},
  pages={8018--8025},
  year={2020}
}

@inproceedings{ren2019generating,
  title={Generating natural language adversarial examples through probability weighted word saliency},
  author={Ren, Shuhuai and Deng, Yihe and He, Kun and Che, Wanxiang},
  booktitle={Proceedings of the 57th annual meeting of the association for computational linguistics},
  pages={1085--1097},
  year={2019}
}

@inproceedings{devlin2019bert,
  title={Bert: Pre-training of deep bidirectional transformers for language understanding},
  author={Devlin, Jacob and Chang, Ming-Wei and Lee, Kenton and Toutanova, Kristina},
  booktitle={Proceedings of the 2019 conference of the North American chapter of the association for computational linguistics: human language technologies, volume 1 (long and short papers)},
  pages={4171--4186},
  year={2019}
}

@article{radford2019language,
  title={Language models are unsupervised multitask learners},
  author={Radford, Alec and Wu, Jeffrey and Child, Rewon and Luan, David and Amodei, Dario and Sutskever, Ilya and others},
  journal={OpenAI blog},
  volume={1},
  number={8},
  pages={9},
  year={2019}
}

@article{vaswani2017attention,
  title={Attention is all you need},
  author={Vaswani, Ashish and Shazeer, Noam and Parmar, Niki and Uszkoreit, Jakob and Jones, Llion and Gomez, Aidan N and Kaiser, {\L}ukasz and Polosukhin, Illia},
  journal={Advances in neural information processing systems},
  volume={30},
  year={2017}
}

@inproceedings{nguyen2023votetrans,
  title={VoteTRANS: Detecting adversarial text without training by voting on hard labels of transformations},
  author={Nguyen-Son, Hoang-Quoc and Hidano, Seira and Fukushima, Kazuhide and Kiyomoto, Shinsaku and Echizen, Isao},
  booktitle={Findings of the Association for Computational Linguistics: ACL 2023},
  pages={5090--5104},
  year={2023}
}

@inproceedings{bao2021defending,
  title={Defending pre-trained language models from adversarial word substitution without performance sacrifice},
  author={Bao, Rongzhou and Wang, Jiayi and Zhao, Hai},
  booktitle={Findings of the association for computational linguistics: ACL-IJCNLP 2021},
  pages={3248--3258},
  year={2021}
}

@inproceedings{maas2011learning,
  title={Learning word vectors for sentiment analysis},
  author={Maas, Andrew and Daly, Raymond E and Pham, Peter T and Huang, Dan and Ng, Andrew Y and Potts, Christopher},
  booktitle={Proceedings of the 49th Annual Meeting of the Association for Computational Linguistics: Human Language Technologies},
  pages={142--150},
  year={2011}
}

@article{meng2021coco,
  title={Coco-lm: Correcting and contrasting text sequences for language model pretraining},
  author={Meng, Yu and Xiong, Chenyan and Bajaj, Payal and Bennett, Paul and Han, Jiawei and Song, Xia and others},
  journal={Advances in Neural Information Processing Systems},
  volume={34},
  pages={23102--23114},
  year={2021}
}

@inproceedings{chi2021xlm,
  title={XLM-E: Cross-lingual language model pre-training via ELECTRA},
  author={Chi, Zewen and Huang, Shaohan and Dong, Li and Ma, Shuming and Zheng, Bo and Singhal, Saksham and Bajaj, Payal and Song, Xia and Mao, Xian-Ling and Huang, He-Yan and others},
  booktitle={Proceedings of the 60th Annual Meeting of the Association for Computational Linguistics (Volume 1: Long Papers)},
  pages={6170--6182},
  year={2022}
}

@inproceedings{fengcodebert,
  title={Codebert: A pre-trained model for programming and natural languages},
  author={Feng, Zhangyin and Guo, Daya and Tang, Duyu and Duan, Nan and Feng, Xiaocheng and Gong, Ming and Shou, Linjun and Qin, Bing and Liu, Ting and Jiang, Daxin and others},
  booktitle={Findings of the association for computational linguistics: EMNLP 2020},
  pages={1536--1547},
  year={2020}
}

@inproceedings{morris2020reevaluating,
  title={Reevaluating Adversarial Examples in Natural Language},
  author={Morris, John and Lifland, Eli and Lanchantin, Jack and Ji, Yangfeng and Qi, Yanjun},
  booktitle={Findings of the Association for Computational Linguistics: EMNLP 2020},
  pages={3829--3839},
  year={2020}
}

@inproceedings{gao2021making,
  title={Making pre-trained language models better few-shot learners},
  author={Gao, Tianyu and Fisch, Adam and Chen, Danqi},
  booktitle={Proceedings of the 59th annual meeting of the association for computational linguistics and the 11th international joint conference on natural language processing (volume 1: long papers)},
  pages={3816--3830},
  year={2021}
}

@inproceedings{schick2021exploiting,
  title={Exploiting cloze-questions for few-shot text classification and natural language inference},
  author={Schick, Timo and Sch{\"u}tze, Hinrich},
  booktitle={Proceedings of the 16th conference of the European chapter of the association for computational linguistics: main volume},
  pages={255--269},
  year={2021}
}

@article{brown2020language,
  title={Language models are few-shot learners},
  author={Brown, Tom and Mann, Benjamin and Ryder, Nick and Subbiah, Melanie and Kaplan, Jared D and Dhariwal, Prafulla and Neelakantan, Arvind and Shyam, Pranav and Sastry, Girish and Askell, Amanda and others},
  journal={Advances in neural information processing systems},
  volume={33},
  pages={1877--1901},
  year={2020}
}

@article{zou2023universal,
  title={Universal and transferable adversarial attacks on aligned language models},
  author={Zou, Andy and Wang, Zifan and Carlini, Nicholas and Nasr, Milad and Kolter, J Zico and Fredrikson, Matt},
  journal={arXiv preprint arXiv:2307.15043},
  year={2023}
}

@inproceedings{liu2023autodan,
  title={AutoDAN: Generating Stealthy Jailbreak Prompts on Aligned Large Language Models},
  author={Liu, Xiaogeng and Xu, Nan and Chen, Muhao and Xiao, Chaowei},
  booktitle={The Twelfth International Conference on Learning Representations}
}

@article{liu2024autodan,
  title={Autodan-turbo: A lifelong agent for strategy self-exploration to jailbreak llms},
  author={Liu, Xiaogeng and Li, Peiran and Suh, Edward and Vorobeychik, Yevgeniy and Mao, Zhuoqing and Jha, Somesh and McDaniel, Patrick and Sun, Huan and Li, Bo and Xiao, Chaowei},
  journal={arXiv preprint arXiv:2410.05295},
  year={2024}
}

@article{lv2025hyper,
  title={Hyper adversarial tuning for boosting adversarial robustness of pretrained large vision transformers},
  author={Lv, Kangtao and Fan, Wenyan and Cao, Huangsen and Tu, Kainan and Xu, Yihuai and Zhang, Zhimeng and Li, Yang and Ding, Xin and Wang, Yongwei},
  journal={Pattern Recognition},
  pages={112158},
  year={2025},
  publisher={Elsevier}
}

@article{zhao2025negatively,
  title={Negatively correlated ensemble against transfer adversarial attacks},
  author={Zhao, Yunce and Huang, Wei and Liu, Wei and Yao, Xin},
  journal={Pattern Recognition},
  volume={161},
  pages={111155},
  year={2025},
  publisher={Elsevier}
}

\newpage

\label{sec:appendix}
\begin{table*}[h]
\centering
\caption{Examples of RTD-Guard method for generating perturbation examples}
\resizebox{0.9\textwidth}{!}{
\begin{tabular}{@{} lc @{}}
\toprule
 \textbf{Example Class} & \textbf{Example} 
  \\
  \midrule
\begin{minipage}{1in}
  Original \\Example\\
  (Sci/Tech-0.97)
\end{minipage}
&  
\begin{minipage}{3.8in}
   E-mail scam targets police chief Wiltshire Police warns about "phishing" after its fraud squad chief was targeted.
\end{minipage}

 \\ \noalign{\smallskip}\hdashline\noalign{\smallskip}
 
 \begin{minipage}{1in}
  Adversarial\\ Example \\
  (World-0.54)
\end{minipage}
&  
\begin{minipage}{3.8in}
E-mail scam targets police chief Wiltshire Police warns about "phishing" after its \replacecolor{hoax battalion leiter} was targeted.
\end{minipage}
 \\\noalign{\smallskip}\hdashline\noalign{\smallskip}

\begin{minipage}{1in}
  \textbf{RTD Example}\\
  (Sci/Tech-0.96)
\end{minipage}
&  
\begin{minipage}{3.8in}
e - mail scam targets police \textcolor[rgb]{0.00,0.00,1.00}{[MASK]} wiltshire police warns about " phishing " after its \textcolor[rgb]{0.00,0.00,1.00}{[MASK] [MASK] [MASK]}ter was targeted.
\end{minipage}
\\

\midrule
\begin{minipage}{1in}
  Original\\ Example\\
  (World-0.99)
\end{minipage}
&  
\begin{minipage}{3.8in}
Schrder adopts Russian orphan Three-year-old Victoria, from St Petersburg, has been living at the Schrders \#39; family home in Hanover in northern Germany for several weeks. 
\end{minipage}

 \\ \noalign{\smallskip}\hdashline\noalign{\smallskip}
 
\begin{minipage}{1in}
  Adversarial\\ Example \\
  (Sci/Tech-0.72)
\end{minipage}
&  
\begin{minipage}{3.8in}
Schrder \replacecolor{enact} Russian orphan Three-year-old \replacecolor{Clockwork}, from St Petersburg, has been living at the Schrders \#39; family home in Hanover in northern Germany for several weeks. 
\end{minipage}
 \\\noalign{\smallskip}\hdashline\noalign{\smallskip}

\begin{minipage}{1in}
  \textbf{RTD Example}\\
  (World-0.99)
\end{minipage}
&  
\begin{minipage}{3.8in}
schrder en \textcolor[rgb]{0.00,0.00,1.00}{[MASK]} russian orphan three - year - old \textcolor[rgb]{0.00,0.00,1.00}{[MASK][MASK]}, from st petersburg, has been living at the schrders \textcolor[rgb]{0.00,0.00,1.00}{[MASK]} 39 \textcolor[rgb]{0.00,0.00,1.00}{[MASK]} family home in hanover in northern germany for several weeks.
\end{minipage}
\\
\midrule
\begin{minipage}{1in}
  Original \\Example\\
  (Sports-0.99)
\end{minipage}
&  
\begin{minipage}{3.8in}
Today in Athens Leontien Zijlaard-van Moorsel of the Netherlands wipes a tear after winning the gold medal in the women \#39;s road cycling individual time trial at the Vouliagmeni Olympic Centre in Athens on Wednesday. 
\end{minipage}

 \\ \noalign{\smallskip}\hdashline\noalign{\smallskip}
 
\begin{minipage}{1in}
  Adversarial\\ Example \\
  (Sci/Tech-0.63)
\end{minipage}
&  
\begin{minipage}{3.8in}
Today in Athens Leontien Zijlaard-van Moorsel of the \replacecolor{Dutch} wipes a tear after \replacecolor{won} the gold \replacecolor{decorating} in the \replacecolor{females} \#39;s \replacecolor{routing bicycles capita}   time \replacecolor{experiments} at the Vouliagmeni Olimpia Clinics in Athens on \replacecolor{Domingos}. 
\end{minipage}
 \\\noalign{\smallskip}\hdashline\noalign{\smallskip}

\begin{minipage}{1in}
  \textbf{RTD Example}\\
  (Sports-0.98)
\end{minipage}
&  
\begin{minipage}{3.8in}
today in athens leontien zijlaard - van moorsel of the dutch wipes a tear after won the gold decorating in the \textcolor[rgb]{0.00,0.00,1.00}{[MASK]} \# 39 ; s routing bicycles \textcolor[rgb]{0.00,0.00,1.00}{[MASK]} time \textcolor[rgb]{0.00,0.00,1.00}{[MASK]} at the vouliagmeni olimpia \textcolor[rgb]{0.00,0.00,1.00}{[MASK]} in athens on \textcolor[rgb]{0.00,0.00,1.00}{[MASK]}s.
\end{minipage}
\\ 

\bottomrule
\end{tabular}}
\label{tab:ED_samples}
\end{table*}

\end{document}